\documentclass[10pt,lettersize,journal]{IEEEtran}
\usepackage{amsmath,amsfonts}
\usepackage{algorithmic}
\usepackage{algorithm}
\usepackage{array}
\usepackage[caption=false,font=normalsize,labelfont=sf,textfont=sf]{subfig}
\usepackage{textcomp}
\usepackage{stfloats}
\usepackage{url}
\usepackage{verbatim}
\usepackage{graphicx}
\usepackage[utf8]{inputenc}
\usepackage{amssymb}
\usepackage{amsmath}
\usepackage{adjustbox}
\usepackage{multirow}
\usepackage{bm}
\usepackage{xcolor}
\usepackage{siunitx}
\usepackage{colortbl}
\usepackage{color, colortbl}
\definecolor{Orange}{rgb}{0.0,0.8,0.1}
\definecolor{red}{rgb}{1.0, 0.0, 0.0}
\definecolor{LightOrange}{rgb}{0.8,0.7,0.1}
\definecolor{green}{rgb}{0.0, 1.0, 0.0}

\usepackage{cite}
\def\BibTeX{{\rm B\kern-.05em{\sc i\kern-.025em b}\kern-.08em
    T\kern-.1667em\lower.7ex\hbox{E}\kern-.125emX}}
\usepackage{balance}

\begin{document}

\title{QUB-PHEO: A Visual-Based Dyadic Multi-View Dataset for Intention Inference in Collaborative Assembly}

\author{Samuel~Adebayo$^{1,2}$,~\IEEEmembership{Member,~IEEE,}
        Se\'{a}n~McLoone$^{1,2}$,~\IEEEmembership{Senior,~IEEE,}
        and Joost~C.~Dessing$^{1,3}$%
\IEEEcompsocitemizethanks{\IEEEcompsocthanksitem $^{1}$Centre for Intelligent Autonomous Manufacturing Systems, Queen's University Belfast, Northern Ireland, United Kingdom.\protect\\
E-mail: \{sadebayo01, s.mcloone\}@qub.ac.uk
\IEEEcompsocthanksitem $^{2}$School of Electronics, Electrical Engineering and Computer Science, Queen's University Belfast, Northern Ireland.\protect\\
\IEEEcompsocthanksitem $^{3}$School of Psychology, Queen's University Belfast, Northern Ireland, United Kingdom.\protect\\
E-mail: j.dessing@qub.ac.uk
}
\thanks{This research is funded by the Engineering and Physical Sciences and Research Council (EPSRC), United Kingdom.}
}

\maketitle

\begin{abstract}
QUB-PHEO introduces a visual-based, dyadic dataset with the potential of advancing human-robot interaction (HRI) research in assembly operations and intention inference. This dataset captures rich multimodal interactions between two participants, one acting as a 'robot surrogate,' across a variety of assembly tasks that are further broken down into 36 distinct subtasks. With rich visual annotations—facial landmarks, gaze, hand movements, object localization, and more—for 70 participants, QUB-PHEO offers two versions: full video data for 50 participants and visual cues for all 70. Designed to improve machine learning models for HRI, QUB-PHEO enables deeper analysis of subtle interaction cues and intentions, promising contributions to the field. The dataset will be available at \url{https://github.com/exponentialR/QUB-PHEO} subject to an End-User License Agreement (EULA).
\end{abstract}


\section{Introduction}\label{sec:introduction}
\subsection{Background and Motivation}
\IEEEPARstart{H}{uman}-Robot Interaction (HRI) is a multidisciplinary field that explores the dynamics of interaction between humans and robots \cite{c1, c2, c3, c24}. This field is critical for applications across healthcare, manufacturing, education, and entertainment, necessitating robots that can intuitively understand and respond to human intentions, needs, and goals. With the advent of Industry 5.0, there is a growing emphasis on human-centric and intuitive human-robot collaboration, especially in complex assembly tasks requiring precise coordination between human workers and robots \cite{c25, c26}. Such collaboration demands a deep understanding of non-verbal communication cues like posture, gestures, and facial expressions, which are pivotal in human-to-human interactions and equally crucial in HRI.

\begin{figure}[ht!]
    \centering
    \includegraphics[width=0.99\columnwidth]{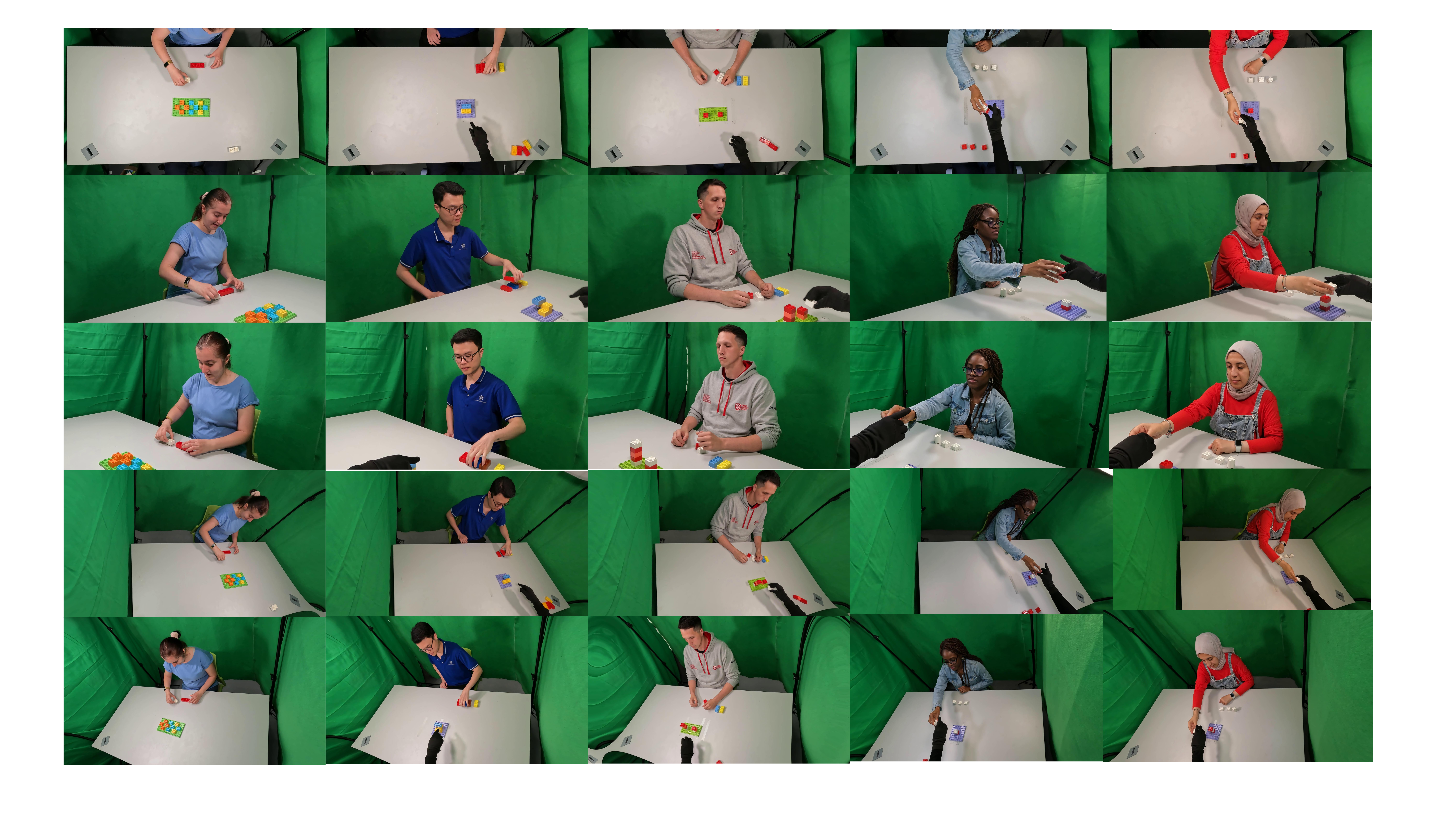}
    \caption{Variety and perspectives in QUB-PHEO: showcasing diverse participant engagement in assembly tasks, captured from multiple camera views to facilitate comprehensive analysis of human-robot interaction.}
    \label{fig:participant_views}
\end{figure}

Effective collaboration in assembly operations hinges on the robot's ability to interpret these non-verbal cues, including hand movements, gaze direction, and object interactions \cite{c27, c28, c30, c31}. Interpreting these cues allows robots to grasp the dynamics of human actions and intentions, leading to safer and more intuitive interactions. For instance, subtle indicators like the direction of gaze or the manner in which an object is manipulated can provide insights into an individual's immediate goals or forthcoming actions. Therefore, the incorporation of these non-verbal visual cues into the robots' perception and decision-making frameworks enhances their ability to interpret the human actions, hence improving predictive capabilities and facilitating smoother collaboration.

However, a major challenge in this domain is the development of systems that can integrate these diverse visual cues to accurately infer human intentions and actions. Although there are a few promising approaches to intention inference \cite{c32, c33, c34, c35, c36}, including our own work \cite{c4}, many often rely on single-view datasets that capture a limited range of visual cues. This limitation can hinder the full understanding of complex human actions, particularly in assembly tasks characterized by the difficulty of hand and object manipulations.

Contrastingly, multi-view datasets offer a more comprehensive visual representation by capturing information from various angles and perspectives. The scarcity of such datasets particularly those encoding subtasks within assembly operations, restricts the development of advanced intention inference methods. Subtask encoding involves annotating the various steps within an assembly task, such as picking up a part, aligning it, or joining parts together.  This limitation constitutes a significant challenge in advancing HRI for complex scenarios. Addressing this gap is crucial for developing advanced systems capable of recognizing and integrating subtle human actions into HRI frameworks. This process provides valuable context that enhances the robot's prediction and understanding of human actions, such as what the next action of the human will be and the corresponding robot's action.

However, many existing approaches are limited by the need to directly simulate or replicate robotic behaviour during experiments, which can constrain the exploration of interaction dynamics \cite{c15, c16, c17, c64, c65}. To overcome this limitation and focus on understanding the foundational aspects of interaction, we introduce the concept of a 'robot surrogate' in assembly operations. \textbf{Robot surrogate} is a human participant who assumes the role of a robot in our experimental setup, unlike approaches that tend to mimic robotic actions, the surrogate's role is to engage in interactions that can uncover the non-verbal communication patterns and dynamics, which can then help to inform the design of robotic bahviours in future research. Hence, utilizing a human surrogate, helps to bypass the current technical limitations of robotic systems to enable a flexible and in-depth exploration of collaborative interactions.

In light of these considerations, we introduce the Perception of Human Engagement in Assembly Operations Vision dataset (QUB-PHEO), a dataset designed specifically for enhancing HRI in assembly operations. QUB-PHEO uniquely focuses on multi-view data collection and the encoding of subtasks, aiming to provide a comprehensive understanding of human action in these scenarios. By capturing a variety of visual cues from multiple perspectives, even without the use of an actual robot, QUB-PHEO seeks to offer a robust foundation for the development of advanced intention inference methods in human-robot collaboration.

\subsection{Contribution}

\begin{figure}[!ht]
    \centering
    \includegraphics[width=3.3in]{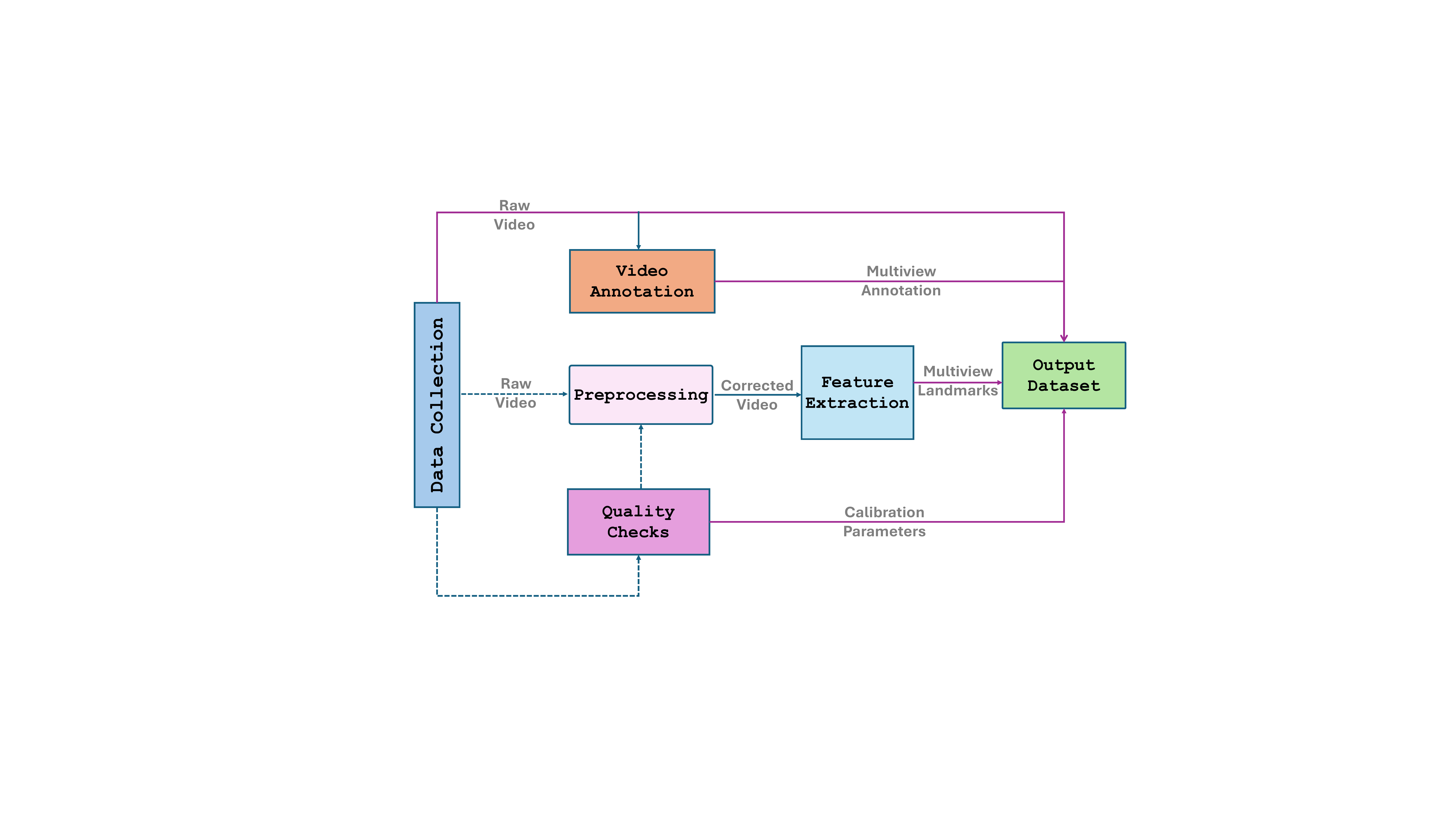}%
    \caption{High-level workflow of the QUB-PHEO dataset development: This flow outlines the systematic approach employed from initial data collection to final dataset generation. Starting with raw video inputs, the workflow incorporates systematic video annotation and rigorous quality checks during preprocessing. The feature extraction stage is then coupled with parallel annotation processes, culminating in the output dataset that integrates corrected video data with feature annotations.}
    \label{fig: PHEO-highlevel}
\end{figure}%
We present the QUB Perception of Human Engagement in Assembly Operations (PHEO) Vision Dataset as an enhancement in Human-Robot Interaction (HRI) research. This dataset leverages insights from human-human interactions to advance the understanding and development of HRI systems. The primary contributions of our work can be divided into direct contributions and broader implications/benefits for the field of HRI:

\subsection*{Direct Contributions}
\begin{itemize}
    \item \textbf{Multi-View Data:} Using a five-camera setup provides different view points essential for understanding interaction dynamics. Our multi-view approach addresses the constraints of single-view datasets by offering a more complete representation of the spatial and temporal aspects of assembly tasks.
    
    \item \textbf{Comprehensive Visual Cues:} QUB-PHEO includes carefully calibrated views and captures a wide range of visual cues, such as hand gestures, body posture, eye gaze, and object manipulation. This rich collection facilitates the extraction of fine-grained keypoints, and by extension, the analysis of non-verbal communication, critical for improving HRI.
    
    \item \textbf{Encoding of Subtasks:} The dataset introduces a classification system for assembly subtasks, featuring 36 distinct categories across several fundamental assembly operations. This granularity enhances the complexity of the dataset and improves the precision of algorithms in interpreting and inferring human intentions within HRI scenarios. 
\end{itemize}

\subsection*{Implications/Benefits for HRI Research}
\begin{itemize}
    \item \textbf{Enhancing Algorithm Development with Comprehensive Data:} The dataset serves as a foundational resource for creating algorithms that foster natural and efficient human-robot collaboration. It supports the development of advanced models adept at interpreting and predicting human actions and intentions with high accuracy. Consequently, it contributes to the evolution of intuitive and symbiotic human-robot partnerships.

    \item \textbf{Addressing Existing HRI Challenges:} By offering multi-dimensional visual data and detailed insights into complex human actions and intentions, QUB-PHEO addresses critical challenges in HRI. It opens avenues for research and innovative applications, aiming to improve cooperation in shared tasks and environments via advanced analytics that facilitate real-time adaptation and mutual understanding between humans and robots.

    \item \textbf{Laying the Groundwork for Practical HRI Applications:} The structured detail within the dataset represents a crucial first step towards developing practical, real-world HRI applications. By providing the necessary context to effectively analyze the sequence and nature of assembly actions, it facilitates the development of sophisticated intention inference algorithms that are essential for creating robust, real-life human-robot interactions outside laboratory settings.

\end{itemize}

\section{Related Work}
Historically, HRI research primarily centered on designing robots for safe and effective collaboration within controlled environments \cite{c37, c38}. This focus laid the groundwork for ensuring robots could perform tasks alongside humans without posing risks, especially in settings like manufacturing or controlled research labs. However, recent advancements have marked a significant shift towards exploring the complexities of dynamic and unstructured interactions. This evolution reflects the broader application of HRI in everyday settings, from domestic to healthcare environments, where interactions are less predictable and more varied \cite{c37, c38, c39}. 

Early applications in teleoperation and telerobotics showcased the importance of human factors in robotics for hazardous environments, highlighting the necessity of effective control, communication, and safety protocols \cite{c37}. The expansion to automated systems, including autonomous vehicles and supervisory control systems, further underscored the critical role of human factors in robot design, particularly to accommodate human capabilities and limitations \cite{c41, c42}. As the domain of HRI has broadened, so too have the challenges, extending into dynamic environments where safety and the ability to understand and anticipate human actions become paramount \cite{c43, c44}. This reiterates the importance of machine learning algorithms and models capable of interpreting complex human behaviors and cues for seamless interaction \cite{c40}.

The transition towards HRI systems for complex assembly tasks illustrates the need for a deeper understanding of human-robot interactions and the importance of seamless communication \cite{c45, c46, c47}. The complexity of interactions in diverse domains highlights the pressing demand for comprehensive datasets\cite{c45, c46, c48}. These resources are crucial for developing interaction models that enable robots to accurately perceive and respond to human cues, aspiring to mirror the complexity and fluidity of human-to-human interactions.

Datasets serve as the backbone of HRI research, enabling the detailed analysis and understanding of human-robot exchanges \cite{c46, c47, c48, c49}. They support the creation of robotic systems attuned to human behavior's variability and task sequences, promoting effective and seamless interactions. Ranging from basic command-response scenarios to complex multimodal engagements, these datasets are instrumental in evolving HRI research.

Despite significant progress, the field continues to face challenges, particularly the need for datasets that capture a wider array of interaction scenarios, including rich multimodal and multi-view data. Such datasets would offer a more complete understanding of interactions, facilitating the development of algorithms that more closely mimic human cognitive and perceptual processes. The call for datasets providing both human and robot perspectives underscores a notable gap, emphasizing the importance of integrating human and social communicative contexts to enhance interaction naturalness and productivity.

In this section, we categorize the discussion into three main areas: General HRI Datasets, Human-Robot Simulators, and Multimodal HRI Datasets, which includes select multi-view datasets. Additionally, it discusses how the QUB-PHEO dataset, through its multi-view data collection and subtask encoding, addresses these gaps and presents new pathways in the study of human-robot collaboration.

\subsection{General HRI Datasets}
The landscape of HRI research is rich with datasets designed to explore various facets of human and robot interplay, varying widely in composition, with a focus on different aspects such as participants involved, types of robots used, the scenarios of interaction, and the data modalities collected, including video, audio, and sensor data. For example, the PINSoRo dataset \cite{c50} offers insights into child-child and child-robot interactions through natural play, emphasizing non-verbal communication in social dynamics. The GoLD dataset \cite{c51} bridges perceptual and linguistic data, focusing on object recognition and language grounding in household environments. The HRI30 dataset \cite{c54} responds to the industrial application gap, with actions tailored to manufacturing and service industries for action recognition and robot learning.  The P2PSTROY dataset \cite{c52} explores peer-to-peer storytelling among children, highlighting the importance of narrative in social interactions and cognitive development. On the other hand, the UE-HRI dataset \cite{c53} captures spontaneous human-robot interactions in public spaces, with a particular focus on engagement and affective states. Collectively, these datasets underscore the complexity of HRI research, showcasing the need for diverse, multimodal datasets to understand and improve the complexities of interactions across various contexts.

\subsection{Human-Robot Interaction Simulators}

\begin{table*}[htbp]
\caption{Summary of HRI Dataset Characteristics}
\label{tab:HRI_datasets}
\centering
\resizebox{\textwidth}{!}{%
\begin{tabular}{llllll}
    \hline
    Dataset & Interaction Setting & Participants & Tasks/Session & Duration & Modality \\
    \hline
    MAHNOB Mimicry \cite{c12} & HHI: 2 & 40 & 54 & 11 hours & Audio, video, depth \\
    USC-CreativiT \cite{c13} & HH[I]2: & 16 & 8 & 8 hours & Audio, video, motion data \\
    MULAI \cite{c14} & HH[I]2: & 26 & 13 & 5.9 hours & Audio, video, motion data, physiological signals \\
    MSP-IMPROV \cite{c21} & HHI: 2 & 12 & 6 & 9 hours & Video, audio \\
    MIT Interview \cite{c22} & HHI: 2 & 69 & 138 & 10.5 hours & Video, audio \\
    CMU Panoptic \cite{c9} & HHI: up to 8 & - & 10 x 3 & 5.5 hours & Video, audio, depth, motion data \\
    MATRICS \cite{c18} & HHI: 4 & 40 & 10 x 3 & 9.2 hours & Video, audio, depth, motion data, eye-tracker, head accelerator \\
    Rakovic et. al \cite{c19} & HHI: 2 & 6 & 6 x 4 & - & Video, motion data, eye-tracker \\
    DAMI-P2C \cite{c20} & HH[I]2: & 68 & 65 & 21.6 hours & Video, audio \\
    Talking with Hands 16.2M \cite{c23} & HH[I]2: & 50 & 50 & 50 hours & Audio, motion data \\
    JESTKOD \cite{c15} & HH[I]2: & 10 & 98 & 4.3 hours & Audio, video, motion data \\
    UDIVA \cite{c16} & HH[I]2: & 147 & 188 x 5 & 90.5 hours & Audio, video, heart rate \\
    M-MS \cite{c17} & HHI: 2 & 21 & 41 + & 16.2 hours & Video, audio, ECG \\
    LISI-HHI \cite{c11} & HHI: 2 & 64 & 32x5 & 8.3 hours & Video, audio, depth, motion data, eye-tracker \\
    \textbf{QUB-PHEO} & HHI: 2 & 70 & 9 & $~$36 hours & Video, gaze, object bbox, hand, upperbody pose, subtasks/intention inference class \\
    
    \hline
\end{tabular}%
}
\end{table*}

Aside from datasets that involve direct human participation, the field of HRI research also benefits from simulations enabled by human simulators \cite{c8}, expanding the scope and feasibility of studies in environments that may be impractical or unsafe for humans. The SIGVerse platform, for example, offers a cloud-based VR environment for multimodal human-robot interaction research, enabling studies in rare or dangerous situations without the associated risks. This approach not only enhances safety and reduces costs but also allows for the replication of highly controlled experimental conditions, which are crucial for cognitive psychological research in HRI.

The WoZ4U interface presents a robust and efficient solution for conducting Wizard-of-Oz (WoZ) experiments \cite{c56}, especially with popular social robots like SoftBank's Pepper. This architecture supports natural language processing, nonverbal behaviors, and navigation functionalities, essential for immersive multimodal HRI research. By enabling detailed control over robot interactions and supporting the recording of visual and auditory data from the robot's perspective, WoZ4U facilitates comprehensive post-analysis of HRI experiments.

Moreover, the development and utilization of these simulators underscore the importance of versatility and scalability in HRI research tools. By providing platforms that are not only flexible in their configuration but also accessible to researchers without requiring extensive programming knowledge or specialized equipment, these simulators pave the way for innovative HRI studies. 

While the diverse datasets and simulation platforms in HRI research have significantly expanded our capabilities to understand and improve human-robot interactions, they come with inherent limitations that need addressing. The challenge of capturing fine-grained interactions and multimodal cues, essential for fully understanding the complexities of HRI, is one area where both real-world datasets and simulations fall short. This limitation is compounded by the lack of multiview data, which restricts analysis to single perspectives and potentially overlooks vital interaction dynamics visible only from specific angles.

Moreover, simulation platforms, despite their utility in offering safe and controlled environments for HRI studies, cannot fully replicate the richness and variability of human behavior. The gap between simulated interactions and actual human behavior might introduce biases or overlook subtle yet critical aspects of interactions, such as emotional nuances or the complexity of spontaneous human actions. The limited ability of both datasets and simulators to capture the full spectrum of multimodal human communication—including verbal cues, body language, facial expressions, and physiological responses—highlights the need for advancements in data collection and simulation technologies.

The transition towards more sophisticated, multimodal datasets is indicative of the field's ambition to capture the multifaceted nature of human-robot interactions in their entirety. It is this ambition that seamlessly connects the exploration of general HRI datasets to the forthcoming discussion on Multiview and Multimodal datasets

\subsection{Multimodal HRI Datasets}
Recent advancements in HRI research are significantly propelled by the integration of multimodal data, such as combining audio, visual, and other sensor data types to enrich the representation of human interactions. Table 1, offers a comparison of existing Multimodal datasets in HRI space. 

Amongst these, the \textbf{CMU Panoptic} dataset \cite{c9} captures a variety of social activities, offering an extensive collection of 3D pose annotations from a massively multiview system- facilitating the observation of  group dynamics and individual social gestures through a setup that incorporates 480 views, over 30 HD cameras, and multiple RGB-D sensors. The \textbf{MULAI dataset} stands out for its inclusion of motion and depth data, which can be invaluable in understanding the spatial aspects of HRI, such as proximity and orientation between humans and robots during interaction. The \textbf{Talking with Hands 16.2M} dataset addresses the crucial aspect of non-verbal communication in HRI, with its large-scale capture of gesture and motion data. This dataset is instrumental in advancing gesture recognition technology, facilitating more natural and intuitive interactions where robots can respond to human gestures in real time. Meanwhile, datasets like \textbf{UDIVA} and \textbf{JESTKOD} provide extensive audio and visual data, enhancing the ability of HRI systems to process and engage in complex verbal and nonverbal communication streams, echoing the dynamic nature of human conversation and interaction. Also, the \textbf{MAHNOB Mimicry} dataset, which includes various signals such as audio, video, depth, and physiological data, is pivotal for emotional state recognition and social signal processing. This dataset allows for the study of nuanced responses and can provide a rich basis for developing empathetic robotic systems that react to human affective states.

In the context of the varied landscape of multimodal HRI datasets, the \textbf{Ego4D consortium} \cite{c10}, a collaborative effort by Facebook AI Research and 13 universities around the world, aimed at pushing the frontier of first-person perception, offers an invaluable collection that, while not solely focused on HRI, encompasses a vast egocentric perspective across a spectrum of daily life activities. The dataset includes a range of annotations such as audio, 3D meshes, eye gaze, and stereo videos, offering researchers an expansive ground for developing and testing AI systems that understand and predict human behaviour from a first-person view. This dataset facilitates unique opportunities for artificial intelligence systems to provide user assistance by understanding and predicting human behaviour from a first-person viewpoint. The insights gleaned from Ego4D's rich, egocentric data can inform the development of assistive AI systems, enabling them to interpret and respond to human actions as they unfold from the individual's perspective. Such a dataset underscores the potential for AI to operate intimately within human environments, anticipating needs and actions in a manner akin to natural human intuition.

Similar to the design intention behind QUB-PHEO, \textbf{LISI-HHI dataset} \cite{c11} provides data on dyadic interactions aimed at robot learning. However, QUB-PHEO distinguishes itself by concentrating on fine-grained purely visual cues to elucidate the dynamics of human interaction within the context of assembly tasks. Additionally, QUB-PHEO also includes fine-grained subtasks. This focus on visual signal processing is important for robots tasked with interpreting and adapting to nuanced human behaviours during collaborative tasks, thereby enhancing the robot's ability to engage in complex, precision-driven activities that are central to industrial and manufacturing settings. 

\section{Methodology}
\subsection{Participant Recruitment and Ethics Approval}
The QUB-PHEO dataset was developed with strict adherence to rigorous ethical standards. The research design, methodology, participant recruitment procedures and consent forms were subject to peer review and approval by the Ethics Committee of the Faculty of Engineering and Physical Sciences, at Queen's University Belfast. 

Recruitment was primarily conducted through university-wide communication. In total 70 participants were recruited, ranging from undergraduates to academic staff. These were composed of individuals from diverse backgrounds, including 30\% of Asian descent, 60\% of European descent, and 10\% of African descent. This diverse composition enhances the dataset's applicability across different demographic groups, ensuring the study's findings are reflective of a wide range of human interactions.

Each participant was briefed about the study, the data collection and handling procedures, and the intention to make the resulting dataset publicly available.  They were then asked to sign a digital participation consent and data release form. Of the 70 participants, 50 agreed to make all their data publicity available, including video data, while  the remainder consented to only anonymised data being made available. 

\subsection{Experimental Setup}
\begin{figure}[ht]
    \centering
    \includegraphics[width=\columnwidth]{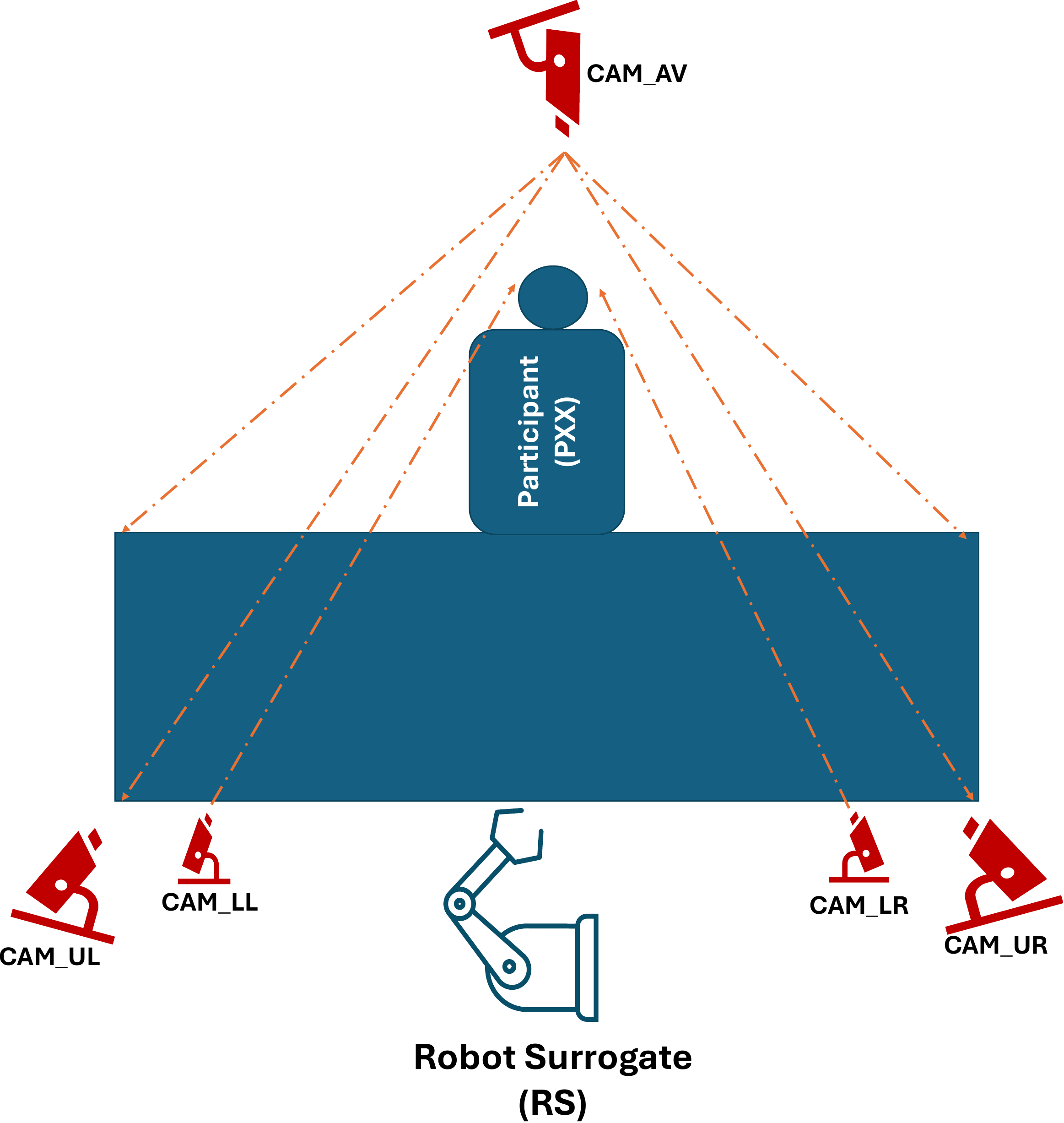}
    \caption{Schematic view of the experimental setup for the QUB-PHEO dataset: This setup depicts a multi-view data collection strategy, employing an aerial view camera ($CAM\_AV$) and four additional cameras at strategic points: lower left ($CAM\_LL$), lower right ($CAM\_LR$), upper left ($CAM\_UL$), and upper right ($CAM\_UR$) to capture the interactions between a human participant ($PXX$) and the robot surrogate ($RS$).}
    \label{fig:Experimental-setup-schematic}
\end{figure}

\begin{figure*}[!ht]
    \centering
    \includegraphics[width=\textwidth]{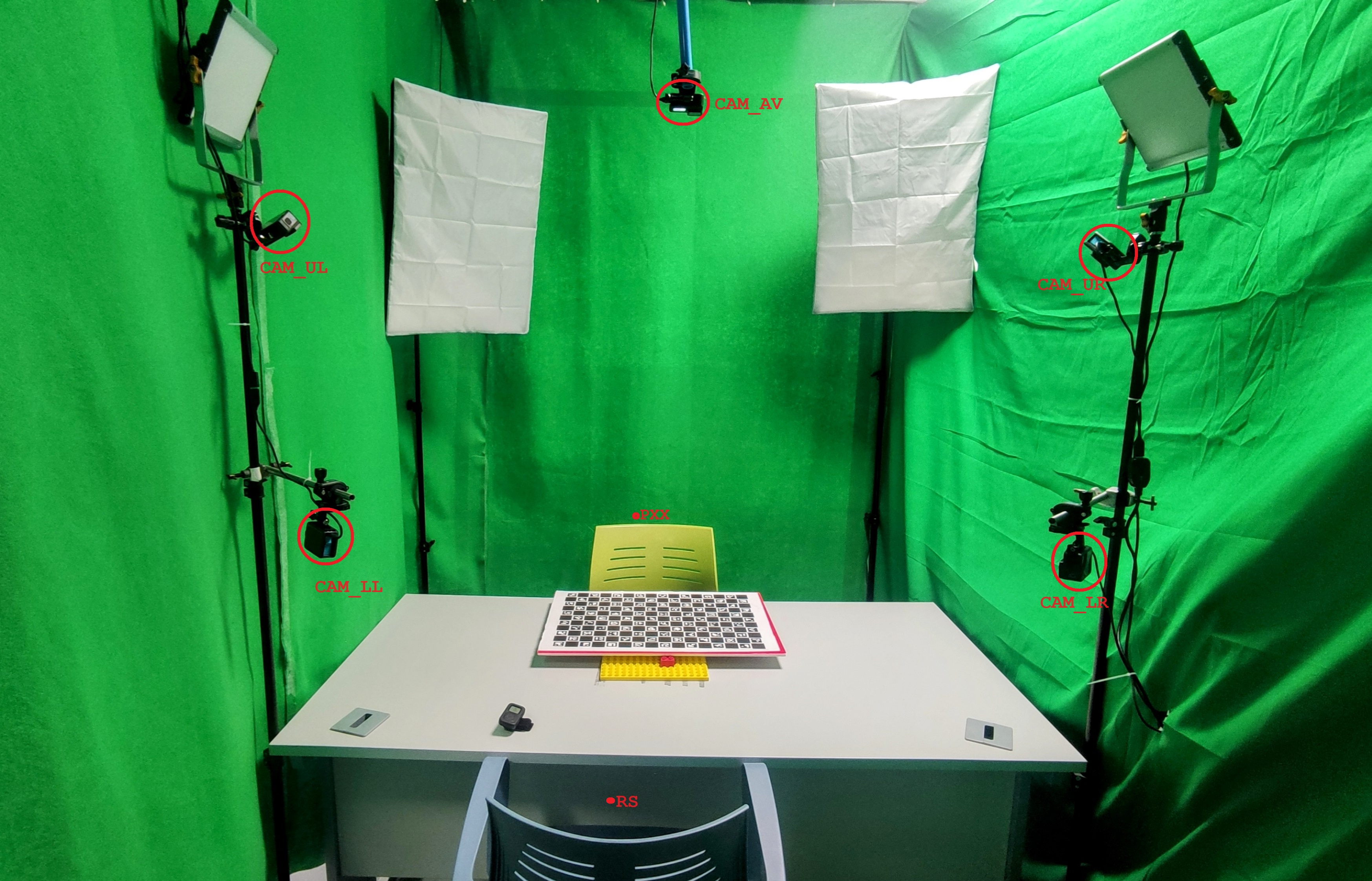}%
    \caption{Pictorial view of our experimental setup for data collection highlighting camera locations: $CAM\_UL$ (upper left), $CAM\_UR$ (upper right), $CAM\_LL$ (lower left), $CAM\_LR$ (lower right) and $CAM\_AV$ (aerial view), the robot surrogate's position ($RS$), and the participant's position ($PXX$).}
    \label{fig: Experimental-setup}
\end{figure*}

The experimental setup was engineered to closely mimic real-world assembly operations, providing a controlled environment that captures the dynamics of human interaction and movement during such tasks. To provide a visual representation of our experimental setup, Figures \ref{fig:Experimental-setup-schematic} and \ref{fig: Experimental-setup}  display the arrangement of our equipment and the environment in which data collection took place. Key Components are labelled to illustrate the configuration of the cameras and the interaction workbench. $CAM\_UL$ and $CAM\_UR$ denote the upper left and upper right cameras respectively, positioned to capture the participant-facing views. '$CAM\_LL$' and '$CAM\_LR$' represent the lower left and lower right cameras offering a side perspective of the interaction space. The '$CAM\_AV$'  camera mounted above the workbench, provides the aerial view of the interactions. $PXX$ typifies the participant's position during data collection while '$RS$' signifies the robot surrogate position.

\textbf{Robot Surrogate Role: } In this study, the robot surrogate role was performed by the attending researcher, allowing for a high degree of control and consistency while maintaining a reasonable level of dynamicity during interaction across all experimental sessions. As the surrogate, the researcher interacted with the participants following standardised protocols (described in the supplementary material) designed to simulate a collaborative assembly task. Also, the surrogate role was not to mimic robotic actions but rather to engage in a way that would help uncover the natural communication patterns and the interaction dynamics essential to HRI.

\begin{itemize}
\item \textbf{Interaction Workbench:} The centrepiece of our setup, a $120$ x $70mm$ table, served as the stage for task execution. This table was chosen for its ample space, accommodating a variety of assembly tasks without restricting participant movement.

\item \textbf{Task Props:} To simulate assembly operations, we employed Duplo blocks by Lego, allowing us to create tasks of varying complexity. These tasks were representative of the subtasks typically found in assembly operations such as sorting, fitting, and stacking.

\item \textbf{Lighting Background:} A professional lighting setup, as shown in Figure \ref{fig: Experimental-setup}, ensured uniform illumination across the interaction space, while a green screen background provided options for background segmentation and uniformity in post-processing.

\item \textbf{Camera Placement and Views:} The setup included five high-definition cameras strategically placed to capture the interaction from multiple angles:
        \begin{enumerate}
            \item \textbf{Aerial-View Camera ($CAM\_AV$):} Provided a top-down view of the workbench, capturing all objects, hand movements, and task progress without occlusion.
            \item \textbf{Participant-Facing Cameras \\($CAM\_LL$, $CAM\_LR$):} Positioned to capture the facial expressions, gaze direction, head pose, and upper-body movement of the participant, essential for analyzing facial cues and gestural communication.
            \item \textbf{Side Cameras ($CAM\_UR$, $CAM\_UL$):} Placed to the sides and slightly elevated, offering a perspective akin to that of a bystander or a robot situated at the workbench, providing a comprehensive view of the interaction space.
        \end{enumerate}
\item \textbf{Participant Arrangement and Choreography:}
        \begin{itemize}
            \item \textbf{Seating Configuration:} The participant and the robot surrogate, whose positions are shown in Figure \ref{fig:Experimental-setup-schematic} as $PXX$ and $RS$, respectively, were seated directly across from each other, mirroring a typical human-robot collaborative setting.
            \item \textbf{Choreography of Interaction:} A set choreography, outlined in the Supplementary material, was followed to standardize the sequence of tasks and interactions. This approach ensured that the dataset reflected consistent interaction patterns across different pairings.
        \end{itemize}
\end{itemize}

\begin{figure*}[!ht]
    \centering
    \includegraphics[width=\textwidth]{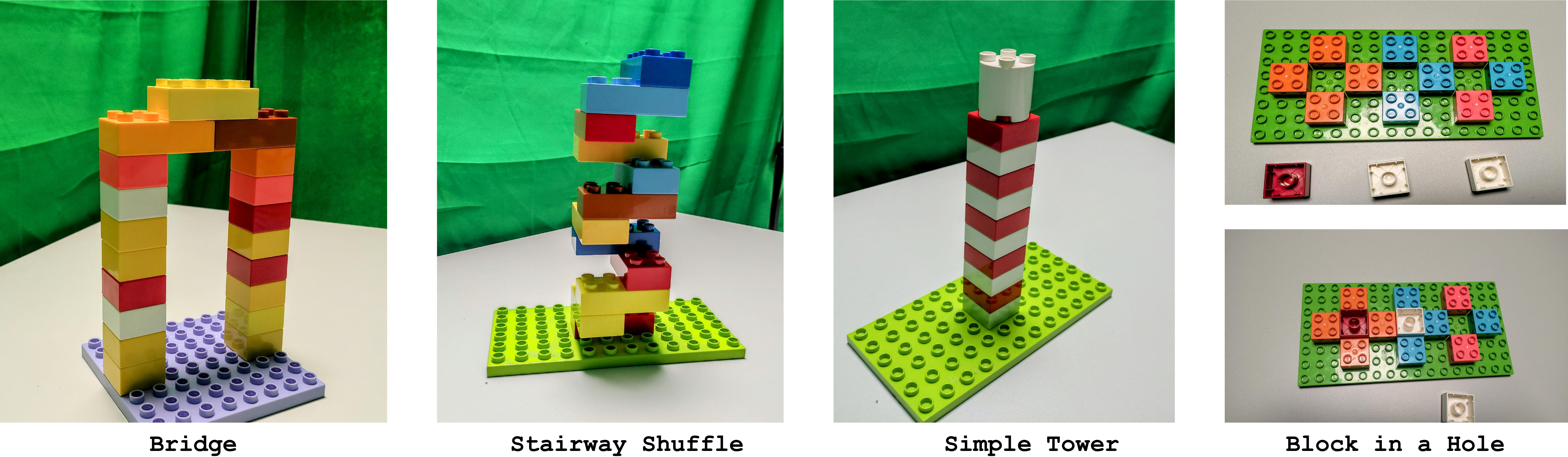}
    \caption{Visual representations of the tasks in the dataset designed to capture a broad range of human interactions: (a) Bridge, (b) Stairway Shuffle, (c) Simple Tower, and (d) Block in a Hole. Each task has variations that challenge different interaction skills.}
    \label{fig:tasks}
\end{figure*}

\subsection{Task Definition}
The tasks in our dataset were designed to replicate assembly operations and capture a detailed sequence of human interactions, complete with subtle cues. Variability in the task design elicits a wide spectrum of interactions. We detail four main tasks, each with at least two variations (unimanual, bi-manual, and collaborative/mid-air stack versions):
 
\begin{enumerate}
    \item \textbf{Block in a Hole (BIAH):} Requiring precise alignment, akin to manufacturing component placement (three variations).
    \item \textbf{Simple Tower (Tower):} Emulating layered construction assembly processes (two variations).
    \item \textbf{Stairway Shuffle:} Simulating sequential ordering common to certain assembly operations (two variations).
    \item \textbf{Bridge Building (Bridge):} Representing tasks demanding both structural understanding and precision, paralleling collaborative engineering projects (two variations).
\end{enumerate}

Selected for their alignment with critical aspects of effective HRI—fine motor skills, spatial reasoning, and sequential problem-solving—the tasks support comprehensive study in this field. Randomization and on-the-fly variations by the robot surrogate (an attending researcher) prevented participant familiarization, thus ensuring a genuine capture of interaction dynamics. Further details on the tasks are provided in the supplementary material. Please see Figure \ref{fig:tasks} for visual representations of these tasks.

\subsection{Subtask Identification}
Actions were classified as subtasks if they constituted a discrete step essential for task progression. For example, in the 'Bridge' task, selecting a support piece was a subtask critical for the overall structure's integrity. Such subtasks, particularly those involving object handover or precise placement, were deemed critical for intention inference, as they signal key decision points in the task flow.
For instance, in the Simple Tower task, subtasks included selecting the correct blocks, placing each block in the correct order, and verifying the tower's stability. Please refer to Section \ref{subsec:annotation} for more detail on subtask annotation.

\label{sec:DataPreprocessing}
\begin{figure*}[!htbp]
    \centering
    \includegraphics[width=\textwidth]{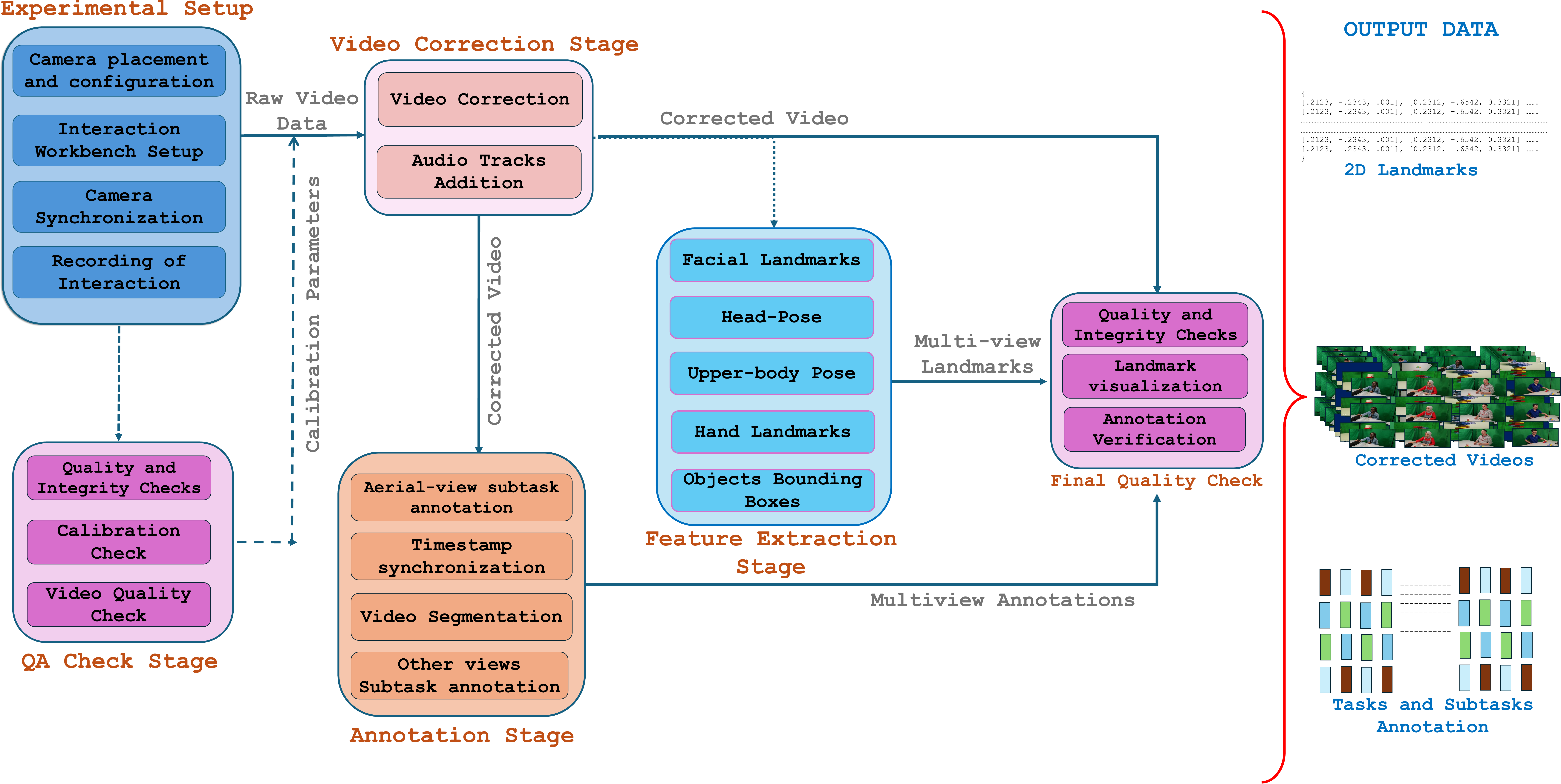}%
    \caption{Methodological pipeline for the data collection, preprocessing, and annotation of the QUB-PHEO dataset}
    \label{fig:PHEO-Pipeline}
\end{figure*}%

\section{Data Collection and Preprocessing}
Figure \ref{fig:PHEO-Pipeline} provides an overview of the methodological pipeline employed in our research. It visually encapsulates the sequential stages from the initial experimental setup to the final data output, underpinned by rigorous quality and integrity checks. The pipeline details the transformation of raw video data into a richly annotated spatially accurate dataset for HRI research. It includes steps for multimodal feature extraction, annotations, landmarks, and segmented videos. Key features include correction of raw video data, extraction of 2D facial landmarks, head-pose, upper-body pose, hand keypoints, and object bounding boxes. We detail the preprocessing of the dataset in subsequent subsections.

\subsection{Calibration and Synchronization} 
\begin{figure}[ht]
    \centering
    \includegraphics[width=\columnwidth]{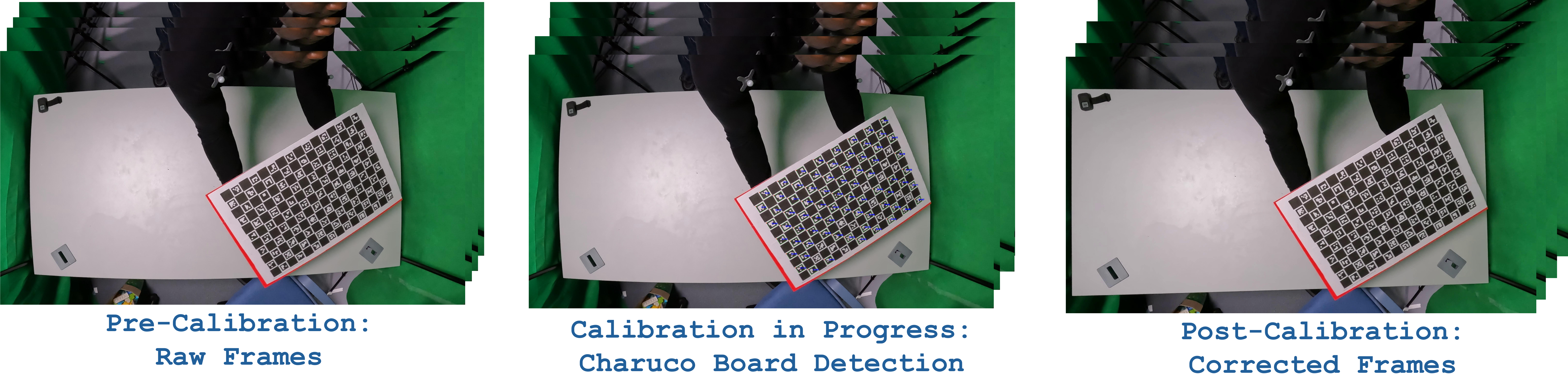}%
    \caption{Calibration Stages of Video Frames. The sequence illustrates the transition from the initial uncorrected state to the final corrected state of a video frame, using a Charuco board for camera calibration. 'Pre-Calibration: Raw Frame' shows the original uncalibrated image, 'Calibration In Progress: Charuco Board Detection' demonstrates the detection of calibration markers, and 'Post-Calibration: Corrected Frame' presents the outcome after applying the calibration process, highlighting the improved accuracy and quality necessary for precise data analysis.}
    \label{fig: Calibration-Charuco}
\end{figure}

Utilizing 5 GoPro Hero10 cameras configured to record at 60fps in 4K Narrow settings, we established temporal synchronization across all cameras using a GoPro remote. This synchronization is pivotal for ensuring the consistency required for multi-angle analysis, stereo calibration, and by extension scene reconstruction, as detailed in Figure \ref{fig:PHEO-Pipeline} under the Calibration and Synchronization Stage. To maintain the highest data fidelity, we implemented a rigorous calibration routine. Our custom-developed calibration tool generates a Charuco board pattern, which, when printed and positioned within the camera's field of view, facilitates the calculation of intrinsic and extrinsic camera parameters.
This calibration process, important for correcting any lens-induced distortions, is exemplified in Figure \ref{fig: Calibration-Charuco}. The sequence begins with the 'Pre-Calibration: Raw Frame,' presenting the unmodified initial video state. During 'Calibration In Progress,' the system identifies the Charuco board's markers, enabling the estimation of camera parameters. The final frame, 'Post-Calibration,' showcases the enhanced clarity and geometric accuracy obtained following the application of calibration corrections. An automated system was developed to confirm the frame rate and assess the quality of the recorded videos, as part of our quality and integrity checks (see Figure \ref{fig:PHEO-Pipeline}, QA Check Stage). Any discrepancies in frame quality, including those related to distortion, were systematically rectified by upscaling the footage through our custom software, thus achieving uniformity across the entire dataset.

Utilizing the calibration data gathered during the experimental setup, we applied corrections for lens distortion and spatial orientation using QubVidCalib\footnote{QubVidCalib\cite{c59} is an open source software developed for camera calibration, video correction, and pattern generation. It outputs both intrinsic and extrinsic parameters, and corrected videos.}. 

\subsection{Gaze Estimation and Mapping} \label{subsec:gaze-mapping}
Leveraging our previous work in gaze direction estimation, including the SLYKLatent model introduced for gaze estimation \cite{c7}, and the development of GazeScape\footnote{GazeScape \cite{c60} is a software package designed for gaze mapping in dyadic interactions.}. Although GazeScape was not directly used in the PHEO pipeline due to the nature of our prerecorded data, the foundational ideas and methodologies influenced our current approach. GazeScape employs a machine learning model to estimate gaze direction from aligned facial images obtained from multiple views (multiview). It combines gaze data from these views (in this case, two views at a time) to produce a unified gaze estimate. This estimate is then mapped onto both the void and aerial views, allowing for a detailed analysis of attention direction and engagement levels throughout the interactions (Figure \ref{fig:PHEO-Pipeline}, Feature Extraction Stage).

\begin{figure}[!h]
    \centering
    \includegraphics[width=\columnwidth]{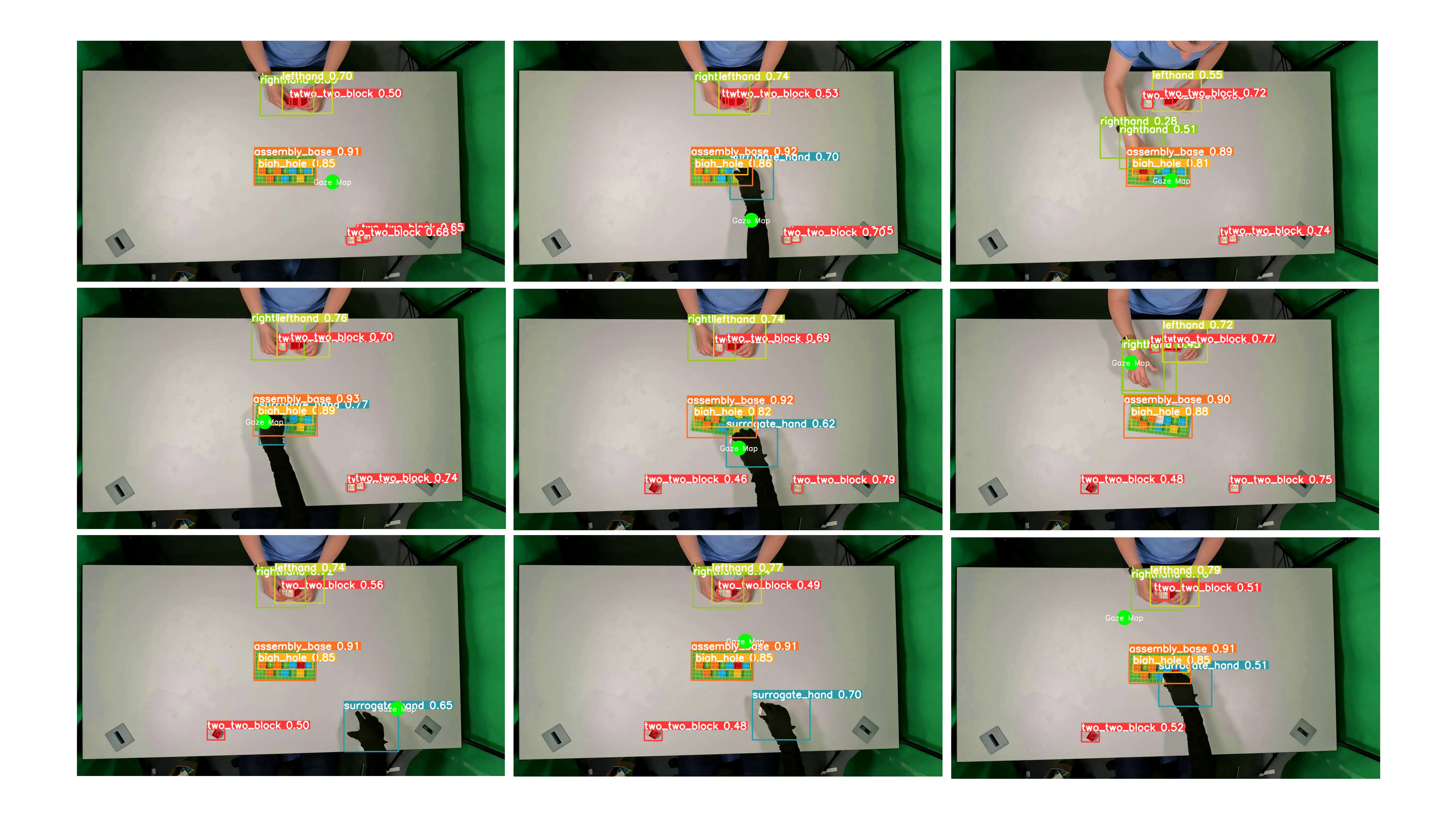}
    \caption{Snapshot sequences of the aerial view with object tracking and gaze mapping (green-filled circle) plots of a participant interacting with the robot surrogate on the BIAH task.}
    \label{fig:participant-aerial-view}
\end{figure}

SLYKLatent requires an aligned facial image with both eyes visible. By leveraging the output of SLYKLatent inference, along with the translation and rotation vectors obtained from the calibration process, we overlaid the translated gaze point estimates on the aerial view. This approach provided a robust understanding of attention direction and engagement levels, contributing valuable insights into non-verbal cues critical for human-robot interaction.

\begin{figure*}[!h]
    \centering
    \includegraphics[width=\textwidth]{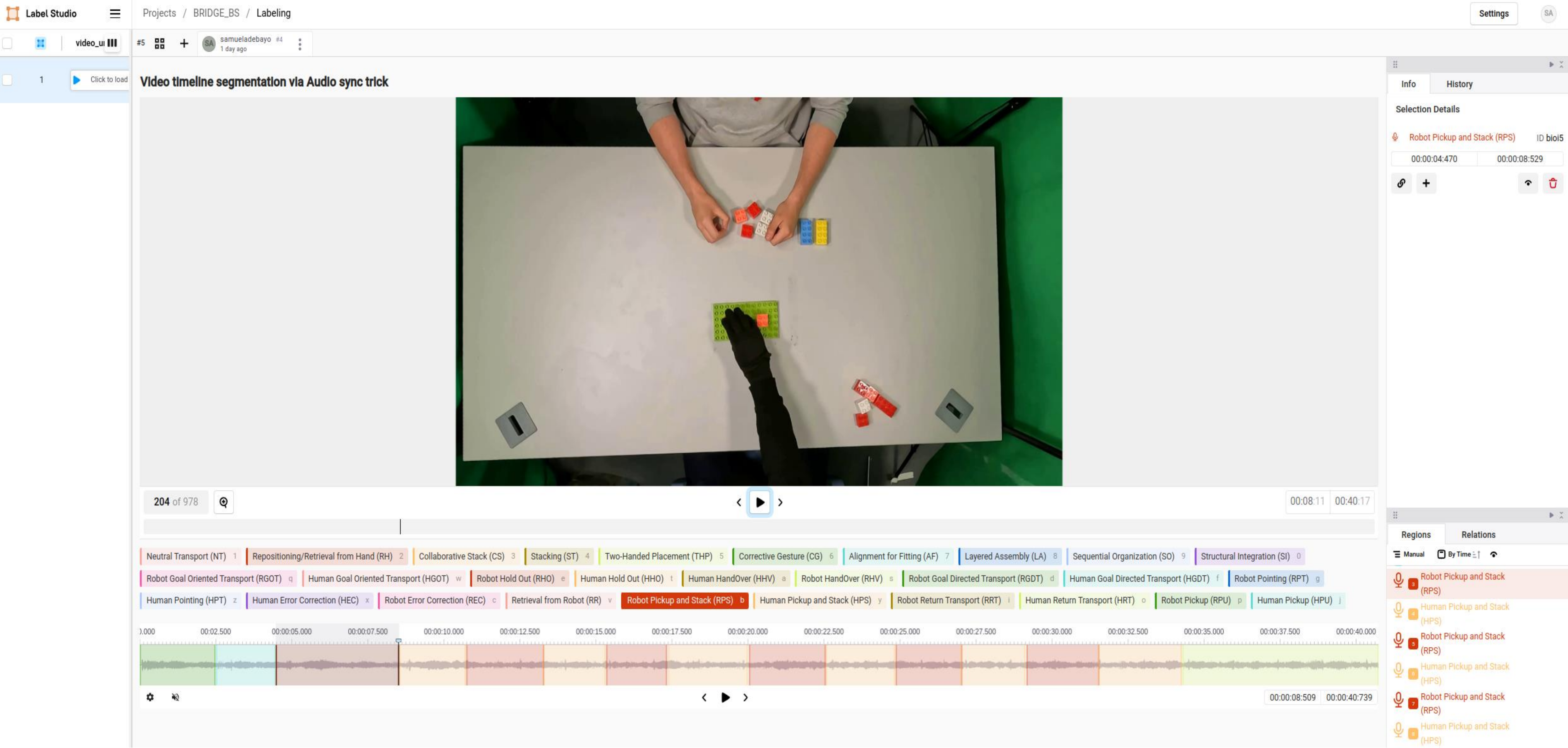}
    \caption{Label Studio interface showing the annotation of assembly tasks. The interface displays various subtask labels and the video timeline, allowing for precise annotation based on synchronized audio and visual cues.}
    \label{fig:label-studio-interface}
\end{figure*}

\subsection{Annotation Process and Subtask Labels} \label{subsec:annotation}
The annotation process for the dataset, as depicted in Figure \ref{fig:label-studio-interface}, was designed to capture the dynamics of dyadic interactions within assembly tasks.  Label Studio \cite{c63}, a versatile tool for video annotations, was employed to achieve detailed and synchronized annotations across multiple camera views. This process was primarily guided by annotations from the aerial perspective.
An essential step in our annotation workflow involves the pre-embedding of audio tracks into the videos. This integration of audio cues is essential as Label Studio leverages these cues for the precise segmentation of video timelines into definable subtasks. Each video segment, delineated by specific audio signals, corresponds to distinct phases of interaction such as 'Robot Pickup and Stack', 'Human Pickup', and 'Collaborative Stack'. These segments are then annotated with start and end timestamps, capturing the subtle cues and interactions observed during the tasks.

Figure \ref{fig:label-studio-interface} illustrates the Label Studio interface used in the annotation process, showcasing the user-friendly environment that facilitated the detailed labeling of interaction phases. The audio-based segmentation allows for highly accurate and temporally precise annotations, enhancing the efficiency and accuracy of our annotation process, which is critical for extensive application in HRI research.

The dataset includes annotations for 36 unique subtasks as presented in Table \ref{tab:subtask-list}, each typically spanning 1 to 7 seconds.

\begin{table}[!ht]
\centering
\caption{Summary of Subtasks/Actions, Their Abbreviations, and Counts}
\begin{tabular}{|c|l|l|r|}
\hline
\textbf{No.} & \textbf{Action} & \textbf{Abbv} & \textbf{Count} \\ \hline
1 & Robot Pickup and Place & RPP & 343 \\ \hline
2 & Human Neutral Swap & HNS & 672 \\ \hline
3 & Human Pickup and Place & HPP & 363 \\ \hline
4 & Robot Neutral Swap & RNS & 228 \\ \hline
5 & Robot Hold Out & RHO & 773 \\ \hline
6 & Collaborative Stack & CS & 396 \\ \hline
7 & Stack Placement & STP & 776 \\ \hline
8 & Human Hold out & HHO & 413 \\ \hline
9 & Human Error Correction & HEC & 15 \\ \hline
10 & Robot Error Correction & REC & 8 \\ \hline
11 & Fallen Over & FO & 14 \\ \hline
12 & Robot Bimanual Pickup and Place & RBPP & 285 \\ \hline
13 & Human Bimanual Neutral Swap & HBNS & 225 \\ \hline
14 & Human Bimanual Pickup and Place & HBPP & 264 \\ \hline
15 & Robot Bimanual Neutral Swap & RBNS & 213 \\ \hline
16 & Human Pickup and Collaborative Stack & HPCS & 369 \\ \hline
17 & Robot Pickup and Collaborative Stack & RPCS & 7 \\ \hline
18 & Robot Bimanual Pickup and Stack & RBPS & 197 \\ \hline
19 & Robot Structural Alignment & RSA & 134 \\ \hline
20 & Human Structural Alignment & HSA & 144 \\ \hline
21 & Human Bimanual Pickup and Stack & HBPS & 220 \\ \hline
22 & Two-Handed Placement & THP & 91 \\ \hline
23 & Robot Pointing & RPT & 591 \\ \hline
24 & Human Pointing & HPT & 542 \\ \hline
25 & Robot Pickup and Stack & RPS & 943 \\ \hline
26 & Human Pickup and Stack & HPS & 964 \\ \hline
27 & Robot Corrective Gesture & RCG & 188 \\ \hline
28 & Human Corrective Gesture & CG & 124 \\ \hline
29 & Base Hold Out & BHO & 118 \\ \hline
30 & Human Goal Directed Transport & HGOT & 458 \\ \hline
31 & Base Position Change & BPC & 413 \\ \hline
32 & Robot Pickup and Handover & RPH & 139 \\ \hline
33 & Human Collection and Place & HCP & 24 \\ \hline
34 & Human Collection and Stack & HCS & 87 \\ \hline
35 & Human Pickup and Handover & HPH & 146 \\ \hline
36 & Robot Collection and Stack & RCS & 145 \\ \hline
\end{tabular}

\label{tab:subtask-list}
\end{table}

The structured approach to video annotation and the detailed subtask labels contribute to the dataset's utility, offering insights into the subtleties of dyadic interaction within assembly tasks.

\begin{figure}[ht]
    \centering
    \includegraphics[width=\columnwidth]{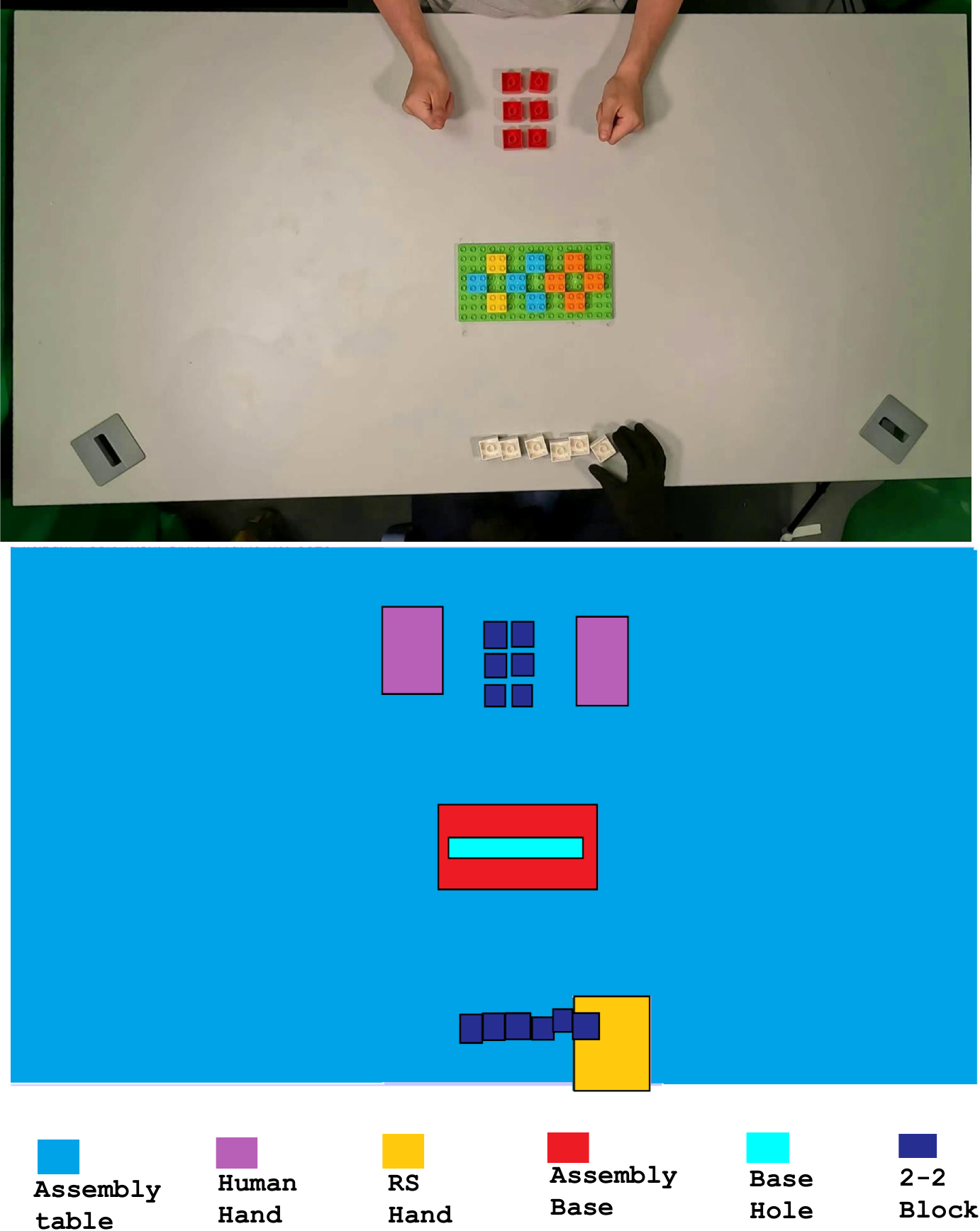}
    \caption{Aerial view of Participant P03 engaged in the Block in a Hole task with 2D bounding box landmark overlays indicating key objects and hands.}
    \label{fig:PHEO-Aerial-view-analysis}
\end{figure}

\subsection{2D Landmark Detection and Interaction Analysis} \label{subsec: 2D-Landmark-Detection-and-Interaction-Analysis}
To capture detailed features and interpret complex interaction dynamics effectively, MediaPipe ML models \cite{c61} were used for detecting hands, upper body pose, and facial landmarks. By extracting 2D landmarks from input frames sourced from CAM\_LL and CAM\_LR cameras, we were able to analyze spatial relationships and movements within the interaction space from a 2D perspective. This methodology is crucial for understanding the subtleties of participant behaviors and interactions, providing insights that are often overlooked in traditional video analysis.

The precision of these 2D landmarks emphasizes the dynamic aspects of human actions and interactions, enriching our dataset with spatially precise data essential for comprehensive behavioral analysis. The extracted landmark data from each frame offers detailed insights into the participant's engagement and interaction with the environment, illustrating the alignment of hand movements and facial expressions with task-specific activities (see Figure \ref{fig:PHEO-Interaction-analysis}, 2D Interaction Analysis). 

To support the academic and research community in replicating and extending our findings, we have included the scripts used for this landmark detection in our data release. Additionally, the 2D landmark data itself is made available, allowing researchers to conduct further analysis or apply the methodology to new datasets.

\subsection{Fine-grained Annotations for Visual Cue Analysis} \label{subsec:fine-grained-annot}
\begin{figure*}[h]
    \centering
    \includegraphics[width=\textwidth]{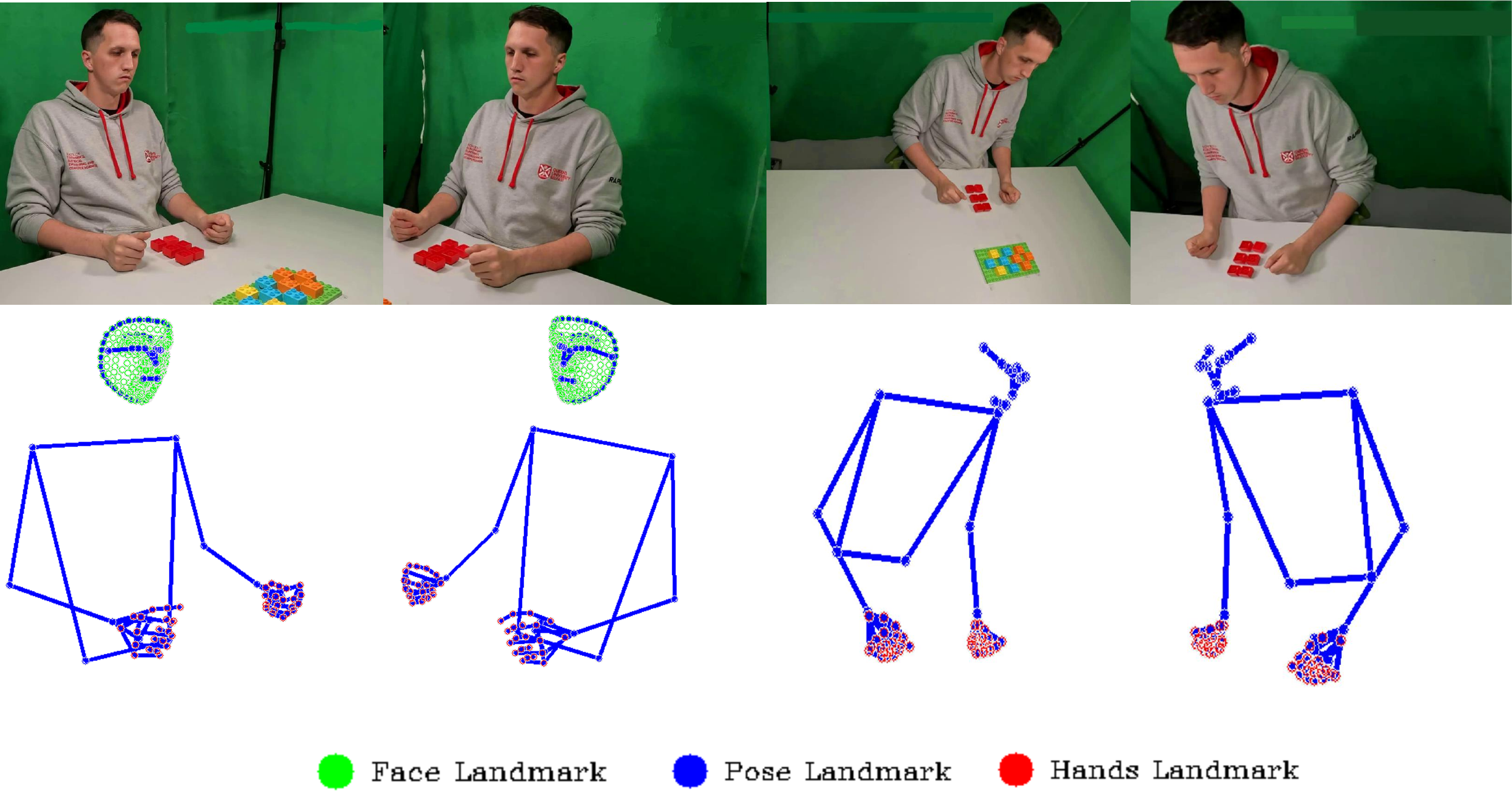}
    \caption{ $CAM_{LL}$, $CAM_{LR}$, $CAM_{UL}$, and $CAM_{UR}$ of Participant P03 engaged in the Block in a Hole task with 2D landmark overlays indicating facial, pose, and hand landmarks.}
    \label{fig:PHEO-Interaction-analysis}
\end{figure*}

Our dataset establishes a new standard for granularity in HRI research through its per-second annotation strategy. Each second of video, comprising $~60$ frames, is annotated to capture a spectrum of visual cues crucial for understanding complex interaction dynamics. These annotations extend beyond the 2D landmarks discussed in Section \ref{subsec: 2D-Landmark-Detection-and-Interaction-Analysis} (see Figure \ref{fig:PHEO-Aerial-view-analysis} and Figure \ref{fig:PHEO-Interaction-analysis}) to include gaze direction, subtask identification, head pose, hand visibility, object bounding boxes, and the specific camera view. This framework provides a detailed snapshot of interaction within each frame, enabling analysis of participant behaviors and interactions.

\begin{figure}[ht]
    \centering
    \includegraphics[width=\columnwidth]{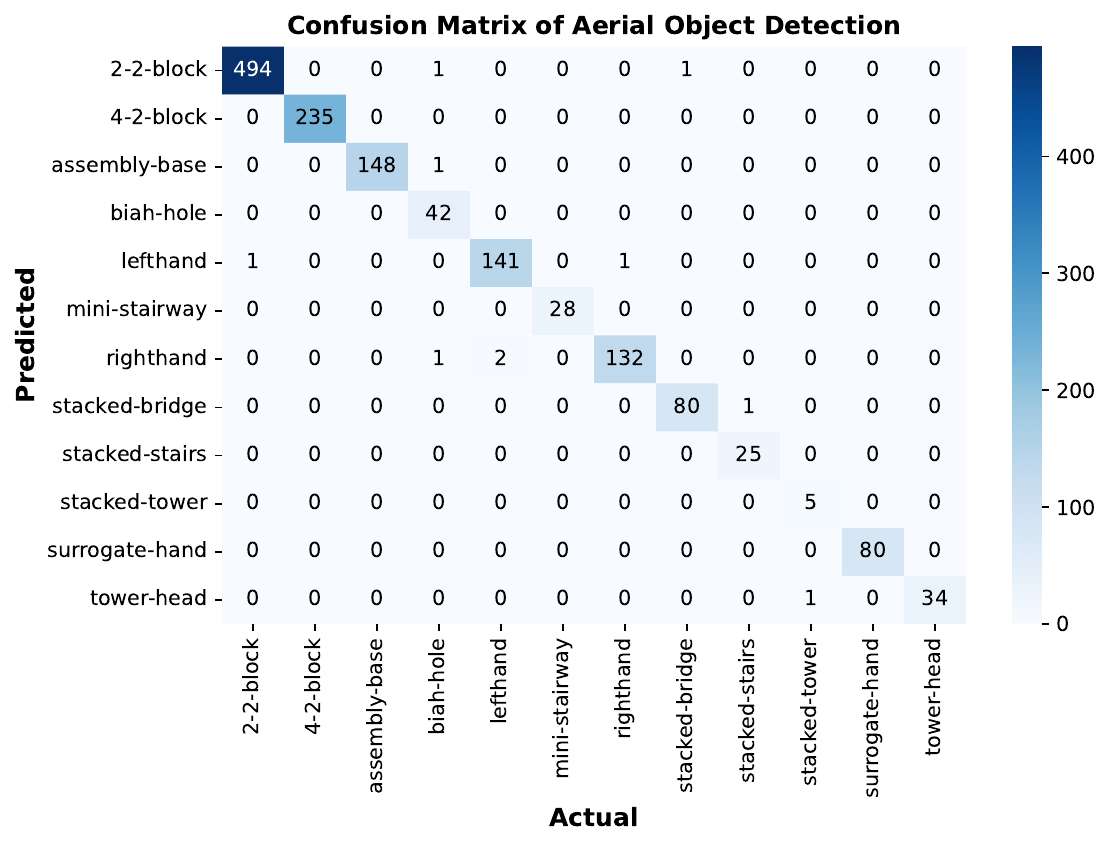}
    \caption{Confusion Matrix result of YOLOV8 model trained via transfer learning for object detection and tracking in the aerial view}.
    \label{fig:confusion-matrix-of-av}
\end{figure}

To further enhance the precision of object recognition within our dataset, a specialized lightweight model based on YOLOV8 was developed and deployed. Ten objects in the scene of interaction were labelled on 1000 images. Using an 80-10-10 dataset split, the YOLOV8 was then trained via transfer learning \cite{c62}, achieving an accuracy of $98\%$ on the test dataset. We then evaluated the model on a live video of $1305$ object instances where it achieved an overall accuracy of $99.2\%$, as summarised in Figure \ref{fig:confusion-matrix-of-av}. This model is particularly adept at recognizing and annotating bounding boxes for key objects and participants in the scene. It focuses on accurately demarcating assembly blocks, human hands, and robot surrogate hands in aerial view videos. The strategic incorporation of this model into our annotation workflow substantially enhances the efficiency and accuracy of the process, reducing the need for manual intervention.

The comprehensive set of the annotations, as summarised in Figure \ref{fig:PHEO-Pipeline} (the Feature Extraction stage),

offers researchers a granular view of HRI, pivotal for crafting algorithms that improve collaborative task performance and understanding. The depth of this dataset facilitates the exploration of complex interaction patterns, advancing the field's capability to design robots that can seamlessly integrate into human environments and collaboratively execute tasks with high levels of proficiency and naturalness.

\subsection{Quality Checks}
At every point during the data collection process rigorous quality checks were performed, as typified in Figure \ref{fig:PHEO-Pipeline}. We carried out three forms of quality checks:

\begin{itemize}
    \item \textbf{Quality and Integrity Checks:} During the experimental setup, the camera placement, configuration, and synchronization were verified for each participant to guarantee the consistency and reliability of the recorded interactions. This stage also involved calibration checks and initial video quality assessments to confirm that the recorded data met our predefined standards.

    \item \textbf{Annotation Stage Quality Control:} During the annotation stage, rigorous quality checks were implemented to ensure the accuracy and robustness of both aerial-view and multi-view annotations. Particular attention was given to the synchronization of timestamps and the segmentation of videos, ensuring that the annotated data accurately reflected the interactions captured. To further guarantee that the correct labels were assigned to time-segmented videos, a software tool named \textbf{aVerify\footnote{https://github.com/exponentialR/aVerify}} was developed. aVerify automates the process of annotation verification by mapping the annotated timestamps to the video, displaying the subtask labels as captions. The user then manually verifies that the captions correspond correctly to the actions in the video by selecting either 'true' or 'false'. If a false label is identified, it can be corrected promptly, ensuring the highest possible accuracy in the dataset.

    \item \textbf{Final Output Quality Assurance:} Before finalizing the dataset, a final quality and integrity check was conducted on the extracted multi-view landmarks and object bounding boxes. This involved verifying the consistency and accuracy of the annotations and performing translational checks across the different views. Only after passing this stringent final quality check were the corrected videos and corresponding annotations included in the output data.

\end{itemize}

These quality checks are critical to maintaining the integrity of the QUB-PHEO dataset, ensuring that the data provided to researchers is both accurate and reliable for subsequent analysis and model training.

\section{Discussion}
\subsection{Descriptive Statistics}
\begin{figure*}[ht]
    \centering
    \includegraphics[width=\textwidth]{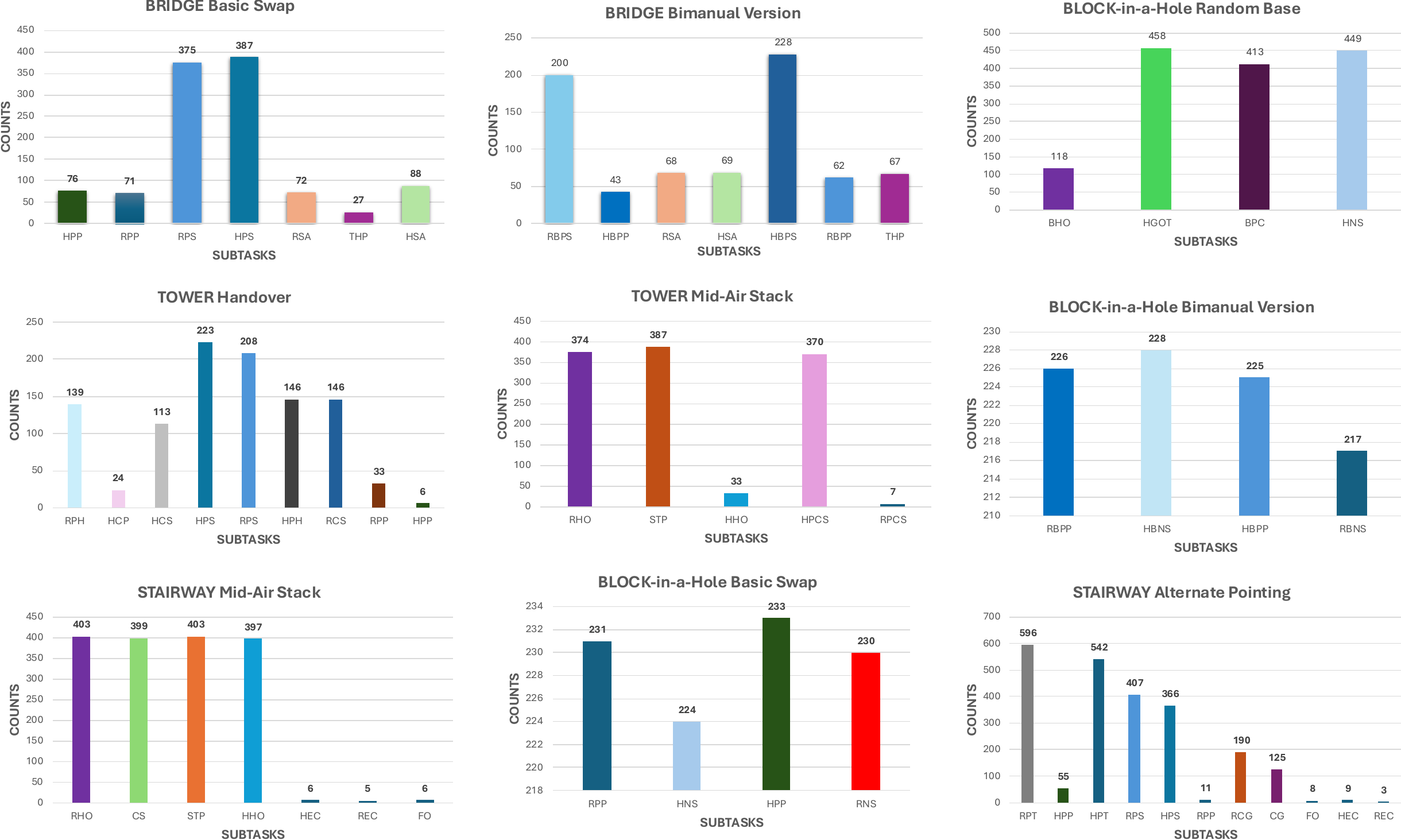}
    \caption{Subtask frequency for each task type: Block in a Hole, Simple Tower, Stairway Shuffle, and Bridge Building.}
    \label{fig:subtask_frequency_by_task_type}
\end{figure*}

Figure \ref{fig:subtask_frequency_by_task_type} presents a grouped bar chart showing the frequency of each subtask within selected task types (Block in a Hole, Simple Tower, Stairway Shuffle, and Bridge Building). This visualization provides insights into the specific demands of different task types, illustrating how various subtasks are utilized to accomplish each task, thereby offering a nuanced understanding of the dataset's structure and the potential areas of research focus.

We commence our analysis by presenting descriptive statistics that encapsulate the PHEO Vision dataset's composition and characteristics. This includes the total number of frames, distribution of subtask labels, durations of interactions, and the variety of visual cues annotated. The objective is to provide an overview of the dataset's scale and the richness of its annotations. Frequency counts of each subtask and visual cue, along with measures such as the mean, median, and range of subtask durations, offer insights into the dataset's diversity and the complexity of captured interactions.

\subsubsection{Dataset Composition}

\begin{table}[ht]
\caption{Summary of High-Level Characteristics of the Dataset}
    \centering
    \begin{tabular}{|l|c|}
        \hline
        \textbf{Characteristic} & \textbf{Value} \\ \hline
        Total Frames & 4.5 Million \\ \hline
        Total Hours of Video & 36 Hours \\ \hline
        Number of Subjects & 70 \\ \hline
        Gender & 31 Male / 39 Female \\ \hline
        Video Data Available for Public & Yes(50)/No(20) \\ \hline
        No Subtasks Labels (videos/Landmarks) & 11032 \\ \hline
    \end{tabular}
    \label{tab:dataset-summary}
\end{table}

Our dataset is rigorously curated to encompass a broad spectrum of dyadic interactions, encapsulating the detailed dynamics essential for in-depth HRI studies. To offer a clear overview of the dataset's scope and diversity, Table \ref{tab:dataset-summary} presents the high-level characteristics, including the number of subjects, demographic distribution, and the availability of video data for public use. 

These metrics underscore the robustness and the analytical depth our dataset offers. As highlighted in Table \ref{tab:dataset-summary}, the dataset comprises a total of \textbf{4.5M} frames, capturing the dynamics of dyadic interaction across \textbf{36} hours of recorded video. For each frame from the side and frontal face cameras (CAM\_LR, CAM\_LL, CAM\_UR, CAM\_UL), there are $478$ face landmarks, $21$ left and right hand landmarks, and $33$ body pose landmarks, all 2D ($x$, $y$).

Additionally, for the aerial view camera, there are $36$ bounding boxes as predicted by the trained object detector described in Section \ref{subsec:fine-grained-annot} and a 2D projected gaze vector ($x$, $y$) as described in Section \ref{subsec:gaze-mapping}. All extracted landmarks are saved in Hdf5.

\subsubsection{Subtask Label Distribution}
Subtask labels are integral to understanding the specific actions and interactions captured within the dataset. There are \textbf{36} unique subtasks labels delineated across various task categories, each representing distinct aspects of assembly and interaction processes. Table \ref{tab:subtasks-details} categorizes the labels into meaningful groups corresponding to their respective tasks, facilitating the analysis of task-specific dynamics, while Figure \ref{fig:subtask_distribution} illustrates the distribution of these subtask labels across all recorded sessions, highlighting the variety and frequency of each action. 

\begin{table}[!ht]
\centering
\caption{Subtask Details for all Tasks Scenarios}
\begin{tabular}{|l|l|}
\hline
\textbf{Task} & \textbf{Subtasks} \\ \hline
\multirow{4}{*}{Block-in-a-Hole Random Base} & BHO \\ \cline{2-2}
                                              & HGOT \\ \cline{2-2}
                                              & BPC \\ \cline{2-2}
                                              & HNS \\ \cline{2-2}
\noalign{\hrule height 1pt} 
\multirow{4}{*}{Block-in-a-Hole-Bimanual Version} & RBPP \\ \cline{2-2}
                                                  & HBNS \\ \cline{2-2}
                                                  & HBPP \\ \cline{2-2}
                                                  & RBNS \\ \cline{2-2}
\noalign{\hrule height 1pt}
\multirow{4}{*}{Block-in-a-Hole Basic Swap} & RPP \\ \cline{2-2}
                                             & HNS \\ \cline{2-2}
                                             & HPP \\ \cline{2-2}
                                             & RNS \\ \cline{2-2}
\noalign{\hrule height 1pt}
\multirow{10}{*}{Tower Handover} & RPH \\ \cline{2-2}
                                 & HCP \\ \cline{2-2}
                                 & HCS \\ \cline{2-2}
                                 & HPS \\ \cline{2-2}
                                 & RPS \\ \cline{2-2}
                                 & HPH \\ \cline{2-2}
                                 & RCS \\ \cline{2-2}
                                 & RCS \\ \cline{2-2}
                                 & RPP \\ \cline{2-2}
                                 & HPP \\ \hline
\noalign{\hrule height 1pt}
\multirow{5}{*}{Tower Mid-Air Stack} & RHO \\ \cline{2-2}
                                      & STP \\ \cline{2-2}
                                      & HHO \\ \cline{2-2}
                                      & HPCS \\ \cline{2-2}
                                      & RPCS \\ \cline{2-2}
\noalign{\hrule height 1pt}
\multirow{7}{*}{BRIDGE Bimanual Version} & RBPS \\ \cline{2-2}
                                         & HBPP \\ \cline{2-2}
                                         & RSA  \\ \cline{2-2}
                                         & HSA  \\ \cline{2-2}
                                         & HBPS \\ \cline{2-2}
                                         & RBPP \\ \cline{2-2}
                                         & THP  \\ \hline
\noalign{\hrule height 1pt} 
\multirow{7}{*}{BRIDGE Basic Setup} & HPP  \\ \cline{2-2}
                                    & RPP  \\ \cline{2-2}
                                    & RPS  \\ \cline{2-2}
                                    & HPS  \\ \cline{2-2}
                                    & RSA  \\ \cline{2-2}
                                    & THP  \\ \cline{2-2}
                                    & HSA  \\
\noalign{\hrule height 1pt}
\multirow{7}{*}{Stairway Mid-Air-Stack} & RHO \\ \cline{2-2}
                                        & HHO \\ \cline{2-2}
                                        & CS  \\ \cline{2-2}
                                        & STP \\ \cline{2-2}
                                        & HEC \\ \cline{2-2}
                                        & REC \\ \cline{2-2}
                                        & FO  \\ \hline
\noalign{\hrule height 1pt} 
\multirow{11}{*}{STAIRWAY Alternate Pointing} & RPT \\ \cline{2-2}
                                        & HPT \\ \cline{2-2}
                                        & HPP \\ \cline{2-2}
                                        & RPS \\ \cline{2-2}
                                        & HPS \\ \cline{2-2}
                                        & RPP \\ \cline{2-2}
                                        & RCG \\ \cline{2-2}
                                        & CG  \\ \cline{2-2}
                                        & FO  \\ \cline{2-2}
                                        & HEC \\ \cline{2-2}
                                        & REC \\ 
\hline
\end{tabular}
\label{tab:subtasks-details}
\end{table}

\subsubsection{Interaction Durations and Visual Cues}
\begin{figure}[ht]
    \centering
    \includegraphics[width=\columnwidth]{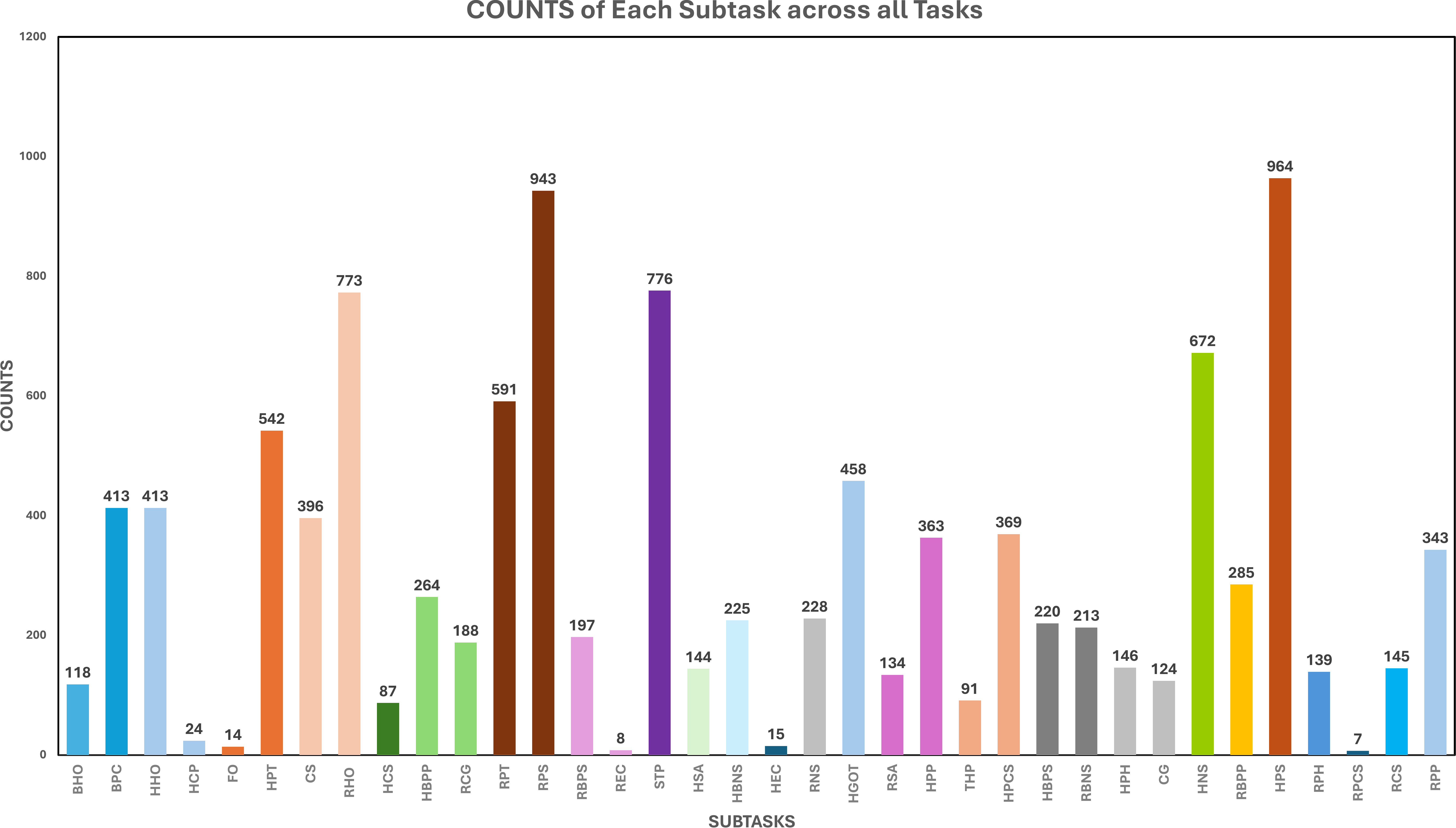}
    \caption{Distribution of subtask labels across tasks. This chart displays the frequency of 36 subtask labels within the QUB-PHEO Vision dataset, aggregated across all tasks and participants. The various colors represent different subtasks, highlighting the diversity and complexity of interactions captured across 70 participants.}
    \label{fig:subtask_distribution}
\end{figure}

Interaction durations and the annotated visual cues offer deeper insights into the complexity of the dataset. The average duration of a subtask is approximately 
4 seconds, with durations ranging from 1 to 7 seconds. The dataset annotates several visual cues, including gaze direction, hand movements, and facial landmarks, to comprehensively capture the subtleties of interaction. The frequency of these cues, detailed in Figure \ref{fig:subtask_frequency_by_task_type} per task level and Figure \ref{fig:subtask_distribution} on a subtask level,  demonstrates the dataset's depth in capturing nuanced interactions.

\subsection{Dataset usage}
QUB-PHEO offers a way for researchers to solve various tasks in HRI using multi-view multi-modal data. To guide researchers in effectively utilising the dataset, we present the mathematical framing of key tasks, along with practical examples of how the dataset can be employed in machine learning models.

\subsubsection{Mathematical Framing of Tasks Using the QUB-PHEO Dataset}
The tasks supported by the PHEO dataset can be mathematically framed to ensure a clear understanding of how to apply the data to specific problems. Below, we describe the mathematical formulation for two primary tasks: subtask classification and intention inference.

\subsubsection*{I. Subtask Classification}
\textbf{Problem Definition: }
The subtask classification task involves assigning each sequence of video frames or extracted keypoints from the dataset to one of the predefined subtask categories. Let \( X = \{x_1, x_2, \ldots, x_T\} \) represent the sequence of video frames or alternatively, a sequence of keypoints extracted from these frames. Here \( T \) is the total number of frames or keypoints, and each \( x_t \) is a feature vector representing the visual information or the corresponding keypoints at time step \( t \).

\textbf{Objective:}
The goal is to learn a function \(f: X \rightarrow y \), where  \( y \) is the subtask label from \( C \) categories. Thus the function can be implemented using models such as convolutional neural networks (CNNs), recurrent neural networks (RNNs) or any form of neural network tailored to process keypoints, with the predicted class given as:

\begin{equation}
    \hat{y} = \arg\max_{c \in \{1, \ldots, C\}} P(y = c \mid X)
\end{equation}

\textbf{Data Preprocessing:} Since subtasks vary in length, padding should be applied to ensure uniform sequence lengths, and masking should be used to ignore padded elements during training to ensure that the model focuses on the relevant parts of each sequence. Keypoints can be normalized or transformed to ensure consistency and effectiveness in classification.

\subsubsection*{II. Intention Inference}
\textbf{Problem Definition:}
The intention inference task involves predicting the next subtask based on either the full or partial sequence of previously observed subtasks, \( Z \). In the former \( Z = \{z_1, z_2, \ldots, z_{t-1}\} \), that is, the entire history of subtasks is included allowing all long-term dependencies to be considered.  In the latter  \( Z = \{z_{t-p}, z_{t-p+1}, \ldots, z_{t-1}\} \), where $p$ denotes the number of previous subtasks considered. This formulation is generally preferred as it is computationally more efficient and  effective when earlier subtasks have less influence on the next prediction.

\textbf{Objective:}
Therefore, the task is to learn a function \(g: Z \rightarrow z_t \), where the next subtask label is predicted by:

\begin{equation}
    \hat{z}_t = \arg\max_{z \in \mathcal{Z}} P(z_t = z \mid Z)
\end{equation}
This can be modelled using sequential models such as RNNs, LSTMs, or Transformers, which are capable of capturing the temporal dependencies and patterns in the sequence of observed subtasks.

\subsubsection*{Loss Functions and Training}
To ensure that the predictions align with the true labels for both tasks, facilitating accurate subtask recognition and reliable intention inference, both subtask classification and intention inference, can use standard categorical cross-entropy loss functions to optimise the models. 

\subsubsection{Tasks Beyond Assembly Operations}
In addition to the above task formulations, QUB-PHEO can be utilised for various recognition tasks involving human images. For instance, the videos of participants seated while performing assembly tasks offer a valuable resource for extracting facial images - which can then be used to train a backbone of self-supervised learning models for facial expression recognition or other facial analysis tasks, even though the dataset does not include explicit ground truth labels for these specific tasks. The same can be said of upper limb recognition tasks.
This makes the QUB-PHEO dataset a valuable resource for both HRI research and broader applications in computer vision and deep learning. Researchers and practitioners can explore these avenues by leveraging the dataset's content which includes millions of images capturing various human assembly operations and activities.

\subsection{Implications for HRI Research}
The QUB-PHEO dataset represents a significant step forward in HRI research, offering unique opportunities to explore complex human interaction patterns in assembly tasks. While the dataset does not include direct human-robot interactions, the use of a human as a robot surrogate to capture interactions from what would be the robot's perspective is a novel approach to understanding how humans might interact with robots in similar contexts. This method allows for the collection of rich data that can inform the development of robotic systems designed to work alongside humans. The following discusses potential applications of the dataset in advancing algorithms for intention inference, task anticipation, and the broader field of HRI, carefully considering the dataset's approach to simulating robot perspectives.

\subsubsection*{Insights from Human Surrogate to Robot Interactions}
Despite the absence of actual robots in the data collection process, the surrogate-based approach provides invaluable insights into the dynamics of human-to-human collaboration that can be extrapolated to human-robot contexts. Specific applications and contributions of the dataset include:

\begin{itemize}
    \item \textbf{Understanding Human-Human Collaboration:} The dataset’s detailed capture of human interactions, seen through the surrogate 'robot' view, sheds light on the subtleties of collaborative behavior, communication, and problem-solving strategies that are likely to be relevant in human-robot teams.
    
    \item \textbf{Informing Robot Design and Behaviour:} Insights gleaned from the dataset can guide the design of robot behaviors that are intuitive and natural for human partners. This includes algorithms for interpreting human actions and intentions, enabling robots to anticipate human needs and respond in supportive ways.
    
    \item \textbf{Training Data for AI Models:} Although the interactions are between humans, the dataset serves as a valuable training resource for AI models focusing on gesture recognition, action prediction, and social dynamics understanding, which are pivotal in enhancing robot autonomy and adaptability in collaborative settings.
    
    \item \textbf{Benchmarking Human-Like Interactions:} The use of a human surrogate allows the dataset to serve as a benchmark for evaluating how closely robot interactions approximate human-like collaboration, aiding in the iterative development of robots that can seamlessly integrate into human workflows.
\end{itemize}

Although the QUB-PHEO dataset does not feature direct human-robot interactions, its approach to capturing dyadic interactions from a robot's vantage point offers a unique lens through which to anticipate and model future human-robot collaboration. This perspective is instrumental in developing robotic assistants that are not only technically adept but are also attuned to the social and communicative cues that facilitate effective teamwork between humans and robots.

\subsection{Accessing and Contributing to Our Dataset}
The extraction of landmarks, gaze, and object detection within our dataset, while robust, is by no means perfect. As with any complex feature extraction process, there are inherent limitations and potential for improvement. We acknowledge that the accuracy of these extracted features may vary depending on several factors, such as lighting conditions, occlusions, and the specific dynamics of interactions captured. To address these challenges and further enhance the quality and utility of the dataset, we invite contributions from the community. Whether it is refining the algorithms for feature extraction, improving landmark accuracy, or enhancing object detection models, your contributions can help to enrich the dataset's analytical depth and applicability.

QUB-PHEO is available online and it is subject to an End-User License Agreement (EULA) \footnote{\url{https://github.com/exponentialR/QUB-PHEO}}.
For those interested in contributing, please refer to our GitHub\footnote{Github Page: https://github.com/exponentialR/QUB-HRI} page where you can find the relevant code, data formats, and guidelines for submission. We welcome collaborations and look forward to working together to advance the field of HRI through improved dataset quality.

\section{Conclusion}
This paper introduced the QUB-PHEO dataset, a novel collection of human interaction data captured through the lens of a robot surrogate, designed to enrich the field of Human-Robot Interaction (HRI) research during collaborative assembly tasks. Through summary statistics, we have showcased the dataset's extensive composition, including a diverse range of subtask labels and visual cues that underpin the complexity and richness of human interactions within assembly tasks. The insights gained from this analysis not only highlight the potential of the QUB-PHEO dataset to deepen our understanding of human collaboration dynamics but also pave the way for the development of more intuitive and effective robotic assistants.

By simulating robot perspectives in human-human interactions, QUB-PHEO offers a unique vantage point for exploring complex interaction patterns that are pivotal for advancing algorithms in intention inference, task anticipation, and adaptive collaboration. Moreover, the dataset serves as a bridge, transferring observational insights from human-human interactions to actionable knowledge that can significantly enhance human-robot collaborations. As such, the QUB-PHEO dataset stands as a contribution to the HRI community, offering a rich resource for researchers and practitioners aiming to explore the nuanced interplay between humans and robots.

\section*{Acknowledgment}
We would like to thank all the volunteers from Queen's University Belfast who took part in our data collection. We would also like to thank Dr Pantelis Sopasakis for providing us with high-performance computing resources to preprocess our dataset.

\begin{IEEEbiography}[{\includegraphics[width=1in,height=1.8in,clip,keepaspectratio]{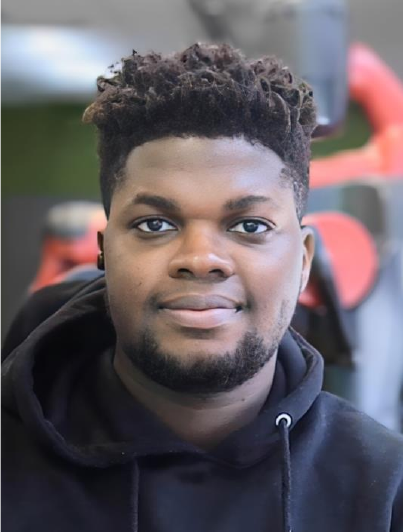}}]{Samuel Adebayo}
earned a First Class BSc in Electronic and Computer Engineering from Lagos State University, Lagos Nigeria (2016-2020) and is currently pursuing his PhD in Machine Learning at Queen's University Belfast, UK (2020-2024). Samuel's research is deeply rooted in computational intelligence, with a primary focus on the intricate dynamics of human hand-eye coordination and its implications for human-robot interactions. This aligns with his broader interest in advancing the fields of Computer Vision, Deep Learning, Natural Language Processing, and Bayesian Machine Learning. Beyond academics, Samuel transitioned from his role as the branch chair of the IEEE Student Branch at Lagos State University to his current leadership position as the Student Branch Chair at Queen's University Belfast.
\end{IEEEbiography}

\begin{IEEEbiography}[{\includegraphics[width=1in,height=1.8in,clip,keepaspectratio]{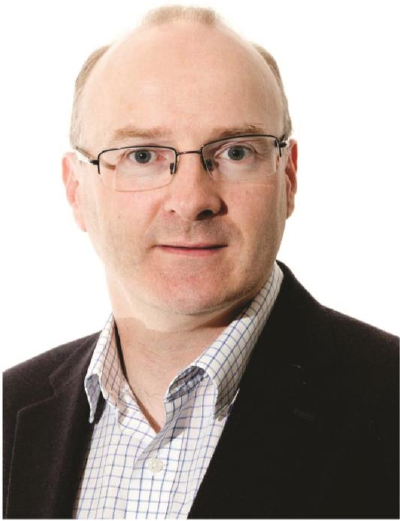}}]
{Sean McLoone} (S$'94$ -- M$'88$  -- SM$'02$) received an M.E. degree in Electrical and Electronic Engineering and a PhD in Control Engineering from Queen's University Belfast, Belfast, U.K. in 1992 and 1996, respectively.
He is currently a Professor and Director of the Energy Power and Intelligent Control Research Centre at Queen's University Belfast. His research interests are in Applied Computational Intelligence and Machine Learning with a particular focus on data based modelling and analysis of dynamical systems, with applications in advanced manufacturing informatics, energy and sustainability, connected health and assisted living technologies.
Prof. McLoone is a Chartered Engineer and Fellow of the Institution of Engineering and Technology. He is a Past Chairman of the UK and Republic of Ireland (UKRI) Section of the IEEE.
\end{IEEEbiography}

\begin{IEEEbiography}[{\includegraphics[width=1in,height=1.8in,clip,keepaspectratio]{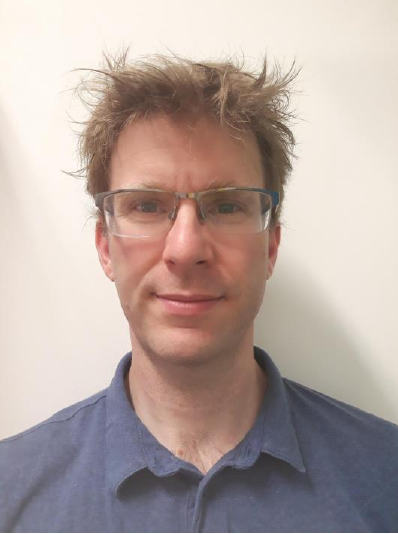}}]
{Joost C. Dessing} received MSc. and PhD degrees in Human Movement Sciences from Vrije University Amsterdam, Amsterdam, The Netherlands, in 2001 and 2005, respectively.
He is currently a Lecturer at the School of Psychology of Queen's University Belfast, where he co-directs the Science in Motion lab. His research interests are fundamental and applied instances of eye-hand coordination, with a particular focus on sports and manufacturing scenarios. He studies these using 3D motion tracking, virtual reality, and neurophysiological tools.
\end{IEEEbiography}
\end{document}